\documentclass{article}

\usepackage[main,final]{neurips_2025}

\usepackage{neurips_2025}
\usepackage{multirow}
\usepackage{multicol}
\usepackage{tabularx}
\usepackage{adjustbox}
\usepackage{bm}
\usepackage[table]{xcolor}

\usepackage{algorithm}
\usepackage[utf8]{inputenc} 
\usepackage[T1]{fontenc}    
\usepackage{hyperref}       
\hypersetup{
  colorlinks,
  linkcolor={red!50!black},
  citecolor={blue!50!black},
  urlcolor={blue!80!black}
  }
\usepackage{url}            
\usepackage{booktabs}       
\usepackage{amsfonts} 
\usepackage{amsmath} 
\usepackage{nicefrac} 
\usepackage{amssymb}
\usepackage{microtype}      
\usepackage{xcolor}         
\usepackage{comment}
\definecolor{orange}{HTML}{ff7f0e}
\definecolor{blue}{HTML}{1f77b4}
\usepackage{nicefrac}
\usepackage{wrapfig}
\usepackage{arydshln}
\usepackage{colortbl}
\definecolor{maroon}{RGB}{128,0,0}
\usepackage{bbm}
\usepackage{enumitem}

\usepackage{cleveref}
\usepackage{empheq}
\usepackage[most]{tcolorbox}
\newtcbox{\mymath}[1][]{%
    nobeforeafter, math upper, tcbox raise base,
    enhanced, colframe=white!30!black,
    colback=white!30, boxrule=1pt,
    #1}
\usepackage{url}
\usepackage{natbib}
\newcommand{\colorA}[1]{\textcolor{burntorange}{#1}}
\newcommand{\colorB}[1]{\textcolor{blue}{#1}}
\newcommand{\colorC}[1]{\textcolor{red}{#1}}

\usepackage{listings}
\usepackage{xcolor}

\definecolor{codegreen}{rgb}{0,0.6,0}
\definecolor{codegray}{rgb}{0.5,0.5,0.5}
\definecolor{codepurple}{rgb}{0.58,0,0.82}
\definecolor{backcolour}{rgb}{0.95,0.95,0.92}

\lstdefinestyle{mystyle}{
    backgroundcolor=\color{backcolour},   
    commentstyle=\color{codegreen},
    keywordstyle=\color{magenta},
    numberstyle=\tiny\color{codegray},
    stringstyle=\color{codepurple},
    basicstyle=\ttfamily\footnotesize,
    breakatwhitespace=false,         
    breaklines=true,                 
    captionpos=b,                    
    keepspaces=true,                 
    numbers=left,                    
    numbersep=5pt,                  
    showspaces=false,                
    showstringspaces=false,
    showtabs=false,                  
    tabsize=2
}

\def\x{\mathbf{x}}

\def\z{\mathbf{z}}

\def\e{\mathbf{e}}

\def\r{\mathbf{r}}

\def\s{\mathbf{s}}

\def\0{\mathbf{0}}

\def\ins{\color{teal}\mathrm{ins}\color{black}}
\def\del{\color{maroon}\mathrm{del}\color{black}}
\def\sub{\color{olive}\mathrm{sub}\color{black}}
\usepackage{wrapfig}
\usepackage{pifont}

\newcommand{\cmark}{\ding{51}} 

\newtheorem{proposition}{Proposition}

\title{Flexible Flows for Biological Sequence Design}

\author{%
  Yogesh Verma\thanks{Now at \href{https://www.openprotein.ai/}{OpenProtein.AI}} \\
  Aalto University, Finland\\
  \texttt{yogesh.verma@aalto.fi} \\
   \And
   Dani Korpela \\
  Aalto University, Finland\\
  \texttt{dani.korpela@aalto.fi} \\
  \AND
  Harri L\"{a}hdesm\"{a}ki \\
  Aalto University, Finland\\
  \texttt{harri.lahdesmaki@aalto.fi} \\
   \And
  Vikas Garg \\
 Aalto University and YaiYai Ltd \\  
\texttt{vgarg@csail.mit.edu} 
}

\begin{document}
\maketitle

\begin{abstract}

Designing functional biological sequences requires navigating vast discrete spaces under strict evolutionary and biophysical constraints. Discrete Flow Matching (DFM) offers a generative framework over such spaces, but existing approaches rely on biologically uninformative couplings and offer limited flexibility for variable length sequence generation and fine-grained control. We propose a structured coupling that encodes domain-specific preferences among sequence elements, biasing the source distribution toward plausible regions without modifying the flow objective or training procedure. Building on this, we introduce a latent edit-based rate parameterization that models variable-length generation via edit operations conditioned on a shared global latent, akin to a latent variable model, while remaining tractable. We further introduce a latent classifier-free guidance mechanism that steers generation coherently in continuous latent space, along with Dirichlet-prior temperature scaling for test-time control over edit operations. Our method achieves state-of-the-art performance across diverse biological sequence tasks, including density estimation, unconditional and conditional DNA sequence generation, and peptide sequence generation. Our code is available at \url{https://github.com/Aalto-QuML/FlexFlow}.

\end{abstract}

\section{Introduction}
\nocite{verma2023abode}
Generative modeling for biological sequences has progressed rapidly across proteins, RNA, and DNA. Early autoregressive approaches such as ProtGPT2 \citep{ferruz2022protgpt2}, ProGen2 \citep{nijkamp2023progen2}, and DNABERT \citep{zhou2023dnabert} demonstrated the viability of language model pretraining for sequence generation, while large-scale protein language models like ESM-2,3 \citep{lin2023evolutionary,hayes2025simulating} and PoET \citep{truong2023poet,truong2025understanding} established powerful representations from evolutionary-scale data. However, autoregressive methods suffer from compounding errors, slow sequential sampling, and limited global coherence, which worsen for long or structurally complex sequences. Discrete diffusion \citep{austin2021structured,shi2024simplified,sahoo2024simple} and flow \citep{gat2024discrete,stark2024dirichlet,tang2025gumbel,davis2024fisher} models have emerged as alternatives, enabling fast, non-autoregressive sampling with improved controllability.

Discrete diffusion \citep{shi2024simplified,austin2021structured} and flow matching \citep{lipman2024flow} 
have demonstrated success in DNA sequence design \citep{stark2024dirichlet,tang2025gumbel,huang2026improving}, protein generation \citep{alamdari2023protein,kong2025protflow}, and recently, in generating peptides as well \citep{tang2025peptune}. Despite these advances, existing discrete methods for biological applications suffer from key limitations: (i) they are typically limited in their ability to generate variable-length sequences, hindering fine-grained control; (ii) they rely on uniform or masked couplings that ignore the structured relationships between biological sequence elements; and (iii) guiding them toward specific functional properties relies on independent, token-wise adjustments \citep{nisonoff2024unlocking} that fail to capture the global dependencies required for complex regulatory or binding motifs.

\textbf{Contributions.} To address these gaps, we propose \textbf{FlexFlow}, a Discrete Flow Matching method for biological sequence design. FlexFlow incorporates a biologically informed structured coupling, an edit-based reverse flow for flexible-length generation with post-hoc operation control, and a latent guidance scheme for fine-grained conditional control. Specifically, our contributions are:
\begin{itemize}
    \item \textbf{Structured Coupling}. We introduce a structured coupling that replaces standard couplings with a domain-specific, biologically informed prior to steer the generative path toward preferred functional neighborhoods. 
    \item \textbf{Latent Edit-based rates with controllable edits}. To support variable-length sequence design, we parameterize rates via edit operations and incorporate Dirichlet priors for post-hoc, temperature-controlled operation mixing, enabling granular control during inference.
    \item \textbf{Latent Guidance}. As an alternative to rate-space classifier-free guidance (CFG), 
    we propose latent guidance that operates entirely within the continuous latent space, ensures global structural consistency, and enhances downstream performance.
    \item \textbf{New Benchmark for peptide sequence generation}. We adapt a novel peptide-MHC II benchmark for generating and evaluating MHC-conditioned peptide sequences.
    \item \textbf{Strong Empirical results}. Our method outperforms existing diffusion and flow baselines in unconditional and conditional DNA sequence generation, as well as peptide generation. 
\end{itemize}
\looseness=-1


\section{Background} 
\label{sec:backrgound}
\subsection{Continuous-time Markov Chains}
Continuous-time Markov Chains (CTMCs) \citep{campbell2024generative,holderrieth2024generator,gat2024discrete,shaul2024flow} are stochastic processes defined over a discrete state space $\mathcal{X}$ that evolve in continuous time, generating trajectories $(X_t)_{t \in [0,1]}$. They are characterized by a rate function $u_t$, which governs the infinitesimal transition probability from state $x_t$ at time $t$ to any other state $x$ at time $t + h$ according to
\begin{align}\label{eq:ctmc}
\mathbb{P}(X_{t+h} = x|X_t = x_t) = \delta_{x_t}(x) + hu_t(x|x_t) + o(h)~,
\end{align}
where $o(h)$ is the contribution from higher order terms such that $\lim_{h \to 0} \frac{o(h)}{h} = 0$. Sampling from a CTMC can be performed by iteratively applying the update rule in \cref{eq:ctmc}. The rate function $u_t(x|x_t)$ specifies the infinitesimal transition probabilities from the current state $x_t$ to any other state $x$ at time $t$. Standard rate conditions $u_t(x|x_t) \geq 0$ for all $x \neq x_t$ and $\sum_x u_t(x|x_t)=0$ ensure that the self-transition rate satisfies $u_t(x_t|x_t) = - \sum_{x \neq x_t} u_t(x|x_t)$.

The rate function $u_t$ induces a probability path $p_t$ if the marginal distribution of the CTMC at each time $t$ coincides with $p_t$, i.e., $X_t \sim p_t$. In particular, these marginals must satisfy the Kolmogorov forward equation, which states that the rate of change of the probability of a state $x$ equals the total incoming infinitesimal probability from other states minus the total outgoing infinitesimal probability from $x$, as determined by the transition rates.

\subsection{Discrete Flow Matching}

Discrete Flow Matching (DFM; \cite{campbell2024generative,gat2024discrete}) provides a simple framework for learning CTMC-based generative models that transport a source distribution $p_0(\x)$ (e.g., noise) to a target distribution $p_1(\x)$ (e.g., data) over a discrete space $\mathcal{X}$. We consider sequences of fixed length $N$, so $\mathcal{X} = \mathcal{T}^{N}$, where $\mathcal{T} = \{1,\ldots,M\}$ is a vocabulary of size $M$. For training, DFM utilizes a conditional CTMC specified by a rate $u_t(\x|\x_t,\x_0,\x_1)$, where samples $(\x_0,\x_1)$ are drawn from a coupling distribution $\pi(\x_0,\x_1)$ with marginals $p_0$ and $p_1$. The rate induces a stochastic process \begin{align*} \label{eq:dfm}
~p_t(\x|\x_0,\x_1), ~~\text{s.t.}~p_0(\x|\x_0,\x_1) = \delta_{\x_0}(\x),~~p_1(\x|\x_0,\x_1) = \delta_{\x_1}(\x)~,
\end{align*}
where $\delta$ denotes the standard Kronecker delta function. Thus, the conditional path interpolates between paired source and target samples, while the learned model follows the corresponding marginal path $p_t(\x)$ that interpolates between the target $p_{\mathrm{data}}(\x) := p_1(\x)$ and the source $p(\x) := p_0(\x)$.
The associated marginal rate, $u_t(\x|\x_t) = \mathbb{E}_{p_t(\x_0,\x_1|\x_t)} u_t(\x|\x_t,\x_0,\x_1)$, defines a CTMC that transports $p_0$ to $p_1$. In practice, models are trained to approximate this rate using objectives such as cross-entropy \citep{gat2024discrete,campbell2024generative} and evidence lower bounds \citep{lou2023discrete,sahoo2024simple,shi2024simplified,shaul2024flow}, which can be viewed within the class of Bregman divergences \citep{holderrieth2024generator}.

\textbf{Token-wise mixture paths.} The specification of the coupling and conditional path is a design choice. Most prior work adopts a factorized, token-wise conditional path of the following form with the rate
\begin{align*}
    p_t(x^i | x^i_0,x^i_1) = (1- \kappa_t) \delta_{x^i_0}(x^i) + \kappa_t \delta_{x^i_1}(x^i), ~u_t(x^i |x^i_t,x^i_0,x^i_1) = \tiny{\frac{\dot{\kappa_t}}{1-\kappa_t} }\left( \delta_{x^i_1}(x^i) - \delta_{x^i_t}(x^i)\right)
\end{align*}
where $\kappa_t$ is a scheduler that satisfies $\kappa_0 = 0, \kappa_1 = 1$. In the multi-dimensional setting, transitions are restricted to single-token changes, yielding 
\vspace{-0.4em}
\begin{align}
    p_t(\x | \x_0,\x_1) = \prod_{i=1}^{N} p_t(x^i | x^i_0,x^i_1),\quad  u_t(\x | \x_t,\x_0,\x_1) = \sum_i \delta_{x_t} (x^{\neg i})u_t(x^i | x^i_t,x^i_0,x^i_1)
\end{align}
where $\delta_{x_t}(x^{\neg i}) = \prod_{k \neq i}\delta_{x_t^k}(x^k) $  indicates that all dimensions except $i$ are held fixed. 
This factorization enables efficient, parallel sampling while requiring only per-dimension parameterization. However, extending beyond token-wise paths is challenging, as more general conditional CTMCs becomes intractable. For more details on CTMCs and DFM, see \citet{lipman2024flow}.
\subsection{Edit Flows}
Edit Flows \citep{havasi2025edit} builds on the DFM framework to define a CTMC-based generative model over variable-length sequences using edit operations while subsuming existing constructions as special cases. The state space is the set of all sequences up to a maximum length $N$, i.e., $\mathcal{X} = \cup_{n=0}^{N}\mathcal{T}^n$. The model parameterizes a time-dependent rate function $u^{\theta}_t$ where transitions $u^{\theta}_t(\x | \x_t)$ are non-zero only if 
$\x$ and $\x_t$ differ by a single edit operation. 

Given a sequence $\x$ of variable length $n(\x)$, the edit operations are mutually exclusive and constitute the support of $u^{\theta}_t(\cdot | \x_t)$. The rates are specified as
\vspace{-0.2em}
\begin{align*}
    \textcolor{teal}{\text{insertion:}} \quad & u^{\theta}_t(\color{teal}\color{teal}\mathrm{ins}\color{black}\color{black}(\mathbf{x},i,a)|\mathbf{x}) = \lambda_{t,i}^{\color{teal}\mathrm{ins} \color{black}}(\x)Q_{t,i}^{\ins}(a|\x) \\
    & \text{with } \color{teal}\mathrm{ins}\color{black}(\x,i,a) = (x_1,\ldots,x_i,a,x_{i+1},\ldots,x_{n(\x)}),\quad \forall i \in \{1,\ldots, n(\mathbf{x}) \}   \\
    \textcolor{maroon}{\text{deletion:}} \quad & u^{\theta}_t(\color{maroon}\mathrm{del}\color{black}(\mathbf{x},i)|\mathbf{x}) = \lambda_{t,i}^{\del}(\x) \\
    & \text{with } \color{maroon}\mathrm{del}\color{black}(\x,i) = (x_1,\ldots,x_{i-1},x_{i+1},\ldots,x_{n(\x)}),\quad \forall i \in \{1,\ldots, n(\mathbf{x}) \} \\
    \textcolor{olive}{\text{substitution:}} \quad & u^{\theta}_t(\color{olive}\mathrm{sub}\color{black}(\mathbf{x},i,a)|\mathbf{x}) = \lambda_{t,i}^{\sub}(\x)Q_{t,i}^{\sub}(a|\x), \\
    & \text{with } \color{olive}\mathrm{sub}\color{black}(\x,i,a) = (x_1,\ldots,x_{i-1},a,x_{i+1},\ldots,x_{n(\x)})  ,\quad \forall i \in \{1,\ldots, n(\mathbf{x}) \}
\end{align*}
where $\lambda_{t,i} \geq 0$ denotes operation-specific rates at position $i$, and $Q_{t,i}$ are normalized distributions over vocabulary for insertion and substitution. This parameterization ensures valid rates, with the self-transition rate given by $u^{\theta}_t(\x_t|\x_t) =  - \sum_{i=1}^{n(\x_t)}\sum_{\mathrm{op}} \lambda_{t,i}^{\mathrm{op}}(\x_t)$, for $\mathrm{op} \in \mathcal{O} = \{\ins, \sub, \del\}$. This formulation enables tractable CTMC modeling directly over sequence space with flexible length changes under the DFM framework. While Edit Flows enables variable-length modeling, it relies on uninformative couplings that ignore the structured relationships between sequence elements and offers no fine-grained control over operations. These limitations carry over to its extensions: EvoFlows \citep{deutschmann2026evoflows} adapts Edit Flows to protein sequence optimization, where both marginals of the flow are natural sequences rather than noise, and inherits the same parameterization unmodified. We instead encode these relationships through a structured coupling, paired with a hierarchical latent parameterization of rates that exposes operation-level control and a novel latent guidance scheme, which we describe next.

\begin{figure}[!t]
    \centering
    \includegraphics[width=\linewidth]{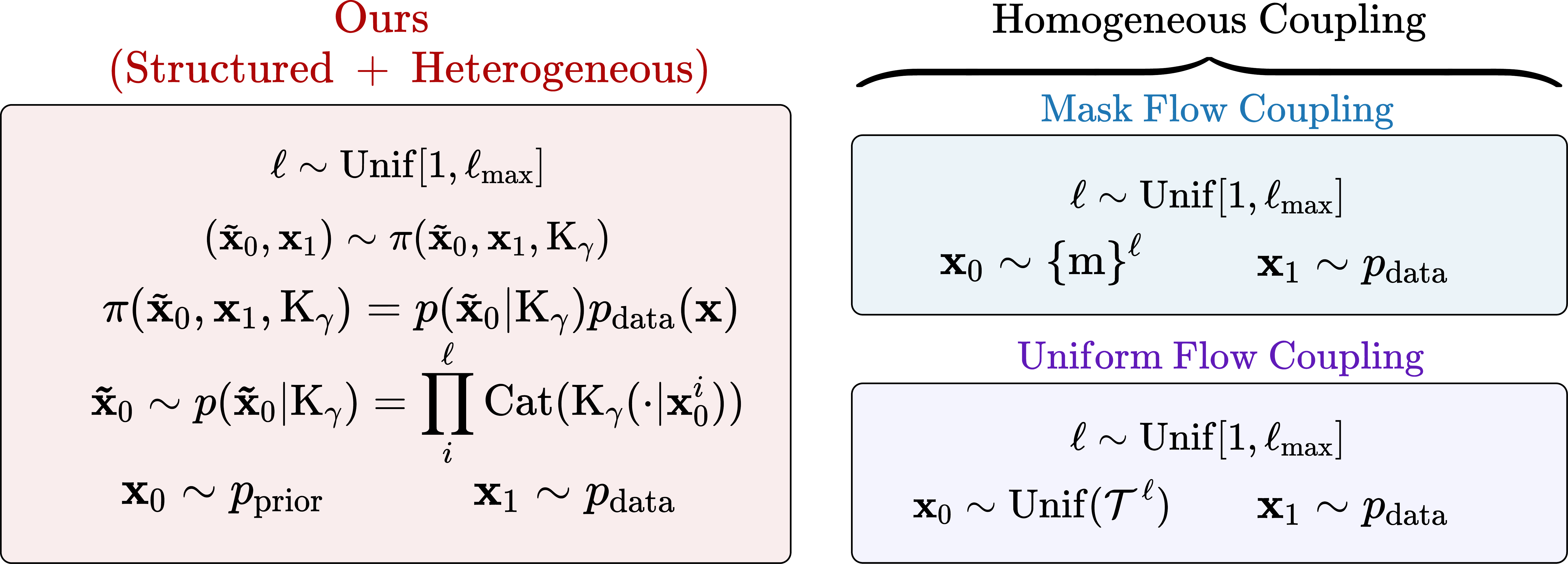}
    \caption{\textbf{Comparison of different couplings}. Description of different couplings used in discrete flow matching and comparison with our proposed structured coupling.}
    \label{fig:couplings}

\end{figure}
\section{Method}
\subsection{Structured Forward Process} 
We aim to impose a structural inductive bias on the forward noising process by tilting it to reflect biologically preferred substitutions, rather than relying on a non-informative baseline. We consider sequences of (fixed) length $N$ over a vocabulary $\mathcal{T}$ of size $M$, so $\mathcal{X} = \mathcal{T}^{N}$ in our setting.

\textbf{Source and target distributions.} The forward process operates over discrete token sequences $\x \in \mathcal{T}^{N}$. Training pairs $(\tilde{\x}_0,\x_1)$ are drawn from a structured coupling $\pi(\tilde{\x}_0,\x_1,\mathrm{K}_\gamma)$ as,
\begin{align}
    (\tilde{\x}_0,\x_1) \sim \pi(\tilde{\x}_0,\x_1,\mathrm{K}_{\gamma}) &= p(\tilde{\x}_0 | \mathrm{K}_\gamma)p_1(\x_1), \quad p(\tilde{\x}_0 | \mathrm{K}_{\gamma}) = \sum_{\x_0} p(\tilde{\x}_0 | \x_0,\mathrm{K}_{\gamma})p_{\mathrm{prior}}(\x_0) \\
    p(\tilde{\x}_0 | \x_0,\mathrm{K}_{\gamma}) &= \prod_{i=1}^{N} \mathrm{Cat}\left(\mathrm{K}_{\gamma}(\cdot|\x^i_0)\right) ,\quad \x_0 \sim p_{\mathrm{prior}}
\end{align}
where $\mathrm{K}_\gamma$ acts as a transition kernel and $\mathrm{K}_{\gamma}(\cdot \mid \x^i_0) = \mathrm{K}_\gamma \e^i_{\x_0}$ gives the conditional distribution over $\mathcal{T}$ corresponding to token $\x^i_0$, 
$\e^i_{\x_0}$ denotes one-hot encoding for the $i^{th}$ token, 
and  $\x_1 \sim p_1 :=p_{\mathrm{data}}$ is a random sample from the training set. The marginal $p(\tilde{\mathbf{x}}_0 | \mathrm{K}_\gamma)$ is obtained by first sampling a sequence $\x_0$ from the prior distribution, and then resampling each token independently from the substitution distribution $\mathrm{K}_\gamma \e^{i}_{\x_0}$ that provides the preferred non-zero substitution probabilities over $\mathcal{T}$.

Here $\gamma$ indexes a family of domain-specific substitution matrices $\{\mathrm{K}_\gamma \in \mathbb{R}^{|\mathcal{T}| \times |\mathcal{T}|}\}$, each encoding preferred mutational substitutions for a given modality: $\mathrm{K}_{\mathrm{prot}}$ for amino acid exchange (e.g., BLOSUM62 \citep{henikoff1992amino}) and $\mathrm{K}_{\mathrm{DNA,RNA}}$ for base substitution biases (e.g., JC69 \citep{jukes1969evolution}, HKY85 \citep{hasegawa1985dating}), see App. \ref{sec:k_gamma} for construction details. The coupling generalizes across sequence taxonomies by simply swapping $\mathrm{K}_\gamma$, leaving the flow and training objective untouched. It also admits a graphical interpretation \citep{weilbach2023graphically,alido2025whitened}, where $\mathrm{K}_\gamma[i,j]$ specifies an edge weight between tokens, inducing a weighted graph whose topology mirrors evolutionary biases. These biases steer the source distribution toward each token's substitution neighborhood rather than the full simplex. With no preference, the coupling reduces to the uniform independent coupling \citep{campbell2024generative,gat2024discrete}, as stated below.

\begin{proposition} \label{prop:coupling}
Let $\mathrm{K}_\gamma = \frac{1}{|\mathcal{T}|}\mathbf{1}\mathbf{1}^\top$, then $p(\tilde{\x}_0 | \mathrm{K}_\gamma) = \mathrm{Uniform}(\mathcal{T}^N)$ and the coupling reduces to the uniform coupling as, $$\pi(\tilde{\mathbf{x}}_0, \mathbf{x}_1) =\mathrm{Uniform}(\mathcal{T}^N)  \,p_{\mathrm{data}}(\mathbf{x})$$
\end{proposition}
The proof is in App.~\ref{sec:proof}, and \Cref{fig:couplings} contrasts our coupling with prior choices. Since $\mathrm{K}_\gamma$ is doubly stochastic, the marginal $p(\tilde{\x}_0 | \mathrm{K}_\gamma)$ stays uniform over $\mathcal{T}$, while the joint $(\x_0, \tilde{\x}_0)$ is correlated according to the substitution preferences in $\mathrm{K}_\gamma$, recovering the uniform independent coupling when $\mathrm{K}_\gamma$ carries no preference.

\textbf{Structured token-wise mixture paths.} Having defined the mutation-tilted coupling $\pi$, we construct a token-wise conditional path \citep{gat2024discrete} between the data sample $\x_1 \sim p_{\mathrm{data}}$ and the tilted endpoint $\tilde{\x}_0 \sim q(\cdot| \mathrm{K_\gamma})$. For each token $i$, the conditional path and rate are defined as
\begin{align*}
    p_t(x^i | \tilde{x}^i_0,x^i_1) = (1- \kappa_t) \delta_{\tilde{x}^i_0}(x^i) + \kappa_t  \delta_{x^i_1}(x^i), ~u_t(x^i |x^i_t,\tilde{x}^i_0,x^i_1) = \tiny{\frac{\dot{\kappa_t}}{1-\kappa_t} }\left( \delta_{x^i_1}(x^i) - \delta_{x^i_t}(x^i)\right)~,
\end{align*}
where $\kappa_t$ is a scheduler that satisfies $\kappa_0 = 0, \kappa_1 = 1$. The multi-dimensional case restricts transitions to those that alter exactly one token at a time,
\begin{align}
    p_t(\x | \tilde{\x}_0,\x_1) = \prod_{i=1}^{N} p_t(x^i | \tilde{\x}^i_0,x^i_1),\quad  u_t(\x | \x_t,\tilde{\x}_0,\x_1) = \sum_i \delta_{x_t} (x^{\neg i})u_t(x^i | x^i_t,\tilde{\x}^i_0,x^i_1)
\end{align}
where $\delta_{\x_t}(x^{\neg i})$  indicates that all dimensions except $i$ are held fixed, and the marginal rate $u_t(\x|\x_t) = \mathbb{E}_{\pi(\tilde{\x}_0,\x_1)} u_t(\x | \x_t,\tilde{\x}_0,\x_1)$ generates the marginal probability path $p_t(\x)$ \citep{gat2024discrete,campbell2024generative}. This factorization restricts transitions to token-wise changes, though sampling across positions \citep{shaul2024flow} at cost of an iterative sampling procedure. The mutation bias enters solely through the tilted endpoints via the coupling $\pi$, biasing transitions toward biologically preferred substitutions, influencing the path and rates.

\subsection{Latent Edit-based rate parameterization}
We take inspiration from \citet{havasi2025edit} to model the reverse process by parameterizing the rate in terms of edit operations as described in section \ref{sec:backrgound}. Rather than directly parameterizing the rate $u^\theta_t$ over the full sequence space, we condition it on a global latent representation $\r \in \mathbb{R}^d$ that encodes sequence-level context and per-token edit operations $\{\mathrm{op}^i\} \in \mathcal{O}^N$, where $\mathcal{O} = \{\ins, \sub, \del\}$. 

Formally, given the current noisy sequence $\mathbf{x}_t$, we first infer the global latent context via:
\begin{align}\label{eq:op_prob}
    p_{\Psi}(\mathbf{r} | \mathbf{x}_t) = \mathcal{N}(\mathbf{r};\, \Psi(\mathbf{x}_t),\, \sigma_r^2 \mathbf{I}), \quad p_{\Phi}(\mathrm{op}^i | x^i_t, \mathbf{r}) = \mathrm{Cat}(\Phi(x^i_t, \mathbf{r}))
\end{align}
where $\Psi(\mathbf{x}_t), \Phi(x^i_t, \mathbf{r})$ are non-linear mappings (e.g., an MLP), and $\sigma_{r}^2$ is the fixed variance. $\Phi(x^i_t, \mathbf{r}) \in \Delta^{2}$ parameterizes the categorical distribution over edit operations at position $i$, conditioned on both the local token state and the global latent context. This hierarchical parameterization induces coupling between token-level operations through the shared latent $\r$, while maintaining conditional independence across positions given $\r$.

Because of this conditional independence, the expected rate $\lambda$ for a specific operation at position $i$ naturally arises by marginalizing over the latent space. Consequently, the full sequence-level marginal rate $u^{\theta}_t(\mathbf{x} | \mathbf{x}_t)$ follows by marginalizing $\z_t$ and summing over per-token edits:
\begin{align}
    u^{\theta}_t(\mathbf{x} | \mathbf{x}_t) =  \mathbb{E}_{\mathbf{r} \sim p_{\Psi}(\mathbf{r} | \mathbf{x}_t)}\left[\sum_{i=1}^{N}\sum_{\mathrm{op} \in \mathcal{O}} p_{\Phi}(\mathrm{op}^i = \mathrm{op} | x^i_t, \mathbf{r})\, u^{\theta}_t\left(\mathrm{op}(\mathbf{x}_t, i,x^i) | \mathbf{x}_t, \mathbf{r}\right)\right]
\end{align}
\subsection{Temperature-controlled Operation Mixing via Dirichlet Priors.}
We can also induce an explicit inherent control over the operation probabilities in eq. \ref{eq:op_prob} by rescaling the predicted logits $\boldsymbol{\pi}_{\Phi} = \Phi(x^i_t, \r) \in \Delta^{3} $ with a position dependent temperature scaling $\boldsymbol{\beta}_i$. This gives the ability to explicitly control the relative rates of insertion, substitution, and deletion operations
during generation as follows:
\begin{align}
    p_{\Phi}(\mathrm{op}^i | x^i_t, \r) &= \mathrm{Cat}(\boldsymbol{\pi}^i_{\Phi} \odot \boldsymbol{\beta}_i), \quad \boldsymbol{\beta}_i = \left[ \frac{1}{\tau^i_{\mathrm{ins}}}, \frac{1}{\tau^i_{\mathrm{sub}}}, \frac{1}{\tau^i_{\mathrm{del}}}\right] \in \mathbb{R}^{3}_{>0}, \quad \sum_{k \in \mathcal{O}} \frac{1}{\tau^i_k} = 1
\end{align}
where $\odot$ denotes element-wise multiplication and $\{\tau^i_k~|~k \in \mathcal{O}\}$ are the per-token per-operation temperature scales. The $\boldsymbol{\beta}_i$ can be modeled via a Dirichlet prior, giving the ability to explicitly control the relative rates of insertion, substitution, and deletion operations during generation as follows:
\begin{align}
    \boldsymbol{\beta}_i = \left[ \frac{1}{\tau^i_{\mathrm{ins}}}, \frac{1}{\tau^i_{\mathrm{sub}}}, \frac{1}{\tau^i_{\mathrm{del}}}\right] \sim \mathrm{Dir}(\boldsymbol{\alpha}), \quad \boldsymbol{\alpha} = [\alpha_{\mathrm{ins}},\alpha_{\mathrm{sub}},\alpha_{\mathrm{del}}] \in \mathbb{R}^{3}_{>0}
\end{align}
where the concentration parameters $\boldsymbol{\alpha}$ act as inverse temperature coefficients, controlling the relative weight of each operation. The concentration parameters $\boldsymbol{\alpha}$ enable test-time control: $\boldsymbol{\alpha}$ can be set post-hoc to steer generation without retraining by fixing the Dirichlet concentration towards any edit operation, according to the desired operation budget.

\subsection{Classifier-free guidance}
We propose two complementary forms of classifier-free guidance for our method, operating at different levels of the model hierarchy: directly on the discrete token rates, and in the continuous global latent space.
\subsubsection{Rate-space Guidance}
A standard way to incorporate conditioning $c$ is classifier-free guidance (CFG) \citep{ho2022classifier}, which combines unconditional and conditional model outputs. This can be applied directly in rate space by interpolating between the unconditional and conditional transition rates \citep{nisonoff2024unlocking,havasi2025edit}. For a guidance strength $w \geq 0$, the guided rate at position $i$ is defined as
\begin{align}
    \tilde{u}^{\theta}_t(x^i | \mathbf{x}_t, \r,c)  &\triangleq u^{\theta}_t(x^i | \mathbf{x}_t, \r)^{1-w}u^{\theta}_t(x^i | \mathbf{x}_t, \r,c)^w  = \hat{\lambda}^{\mathrm{op}}_{t,i}(\x_t,\r,c) \tilde{Q}^{\mathrm{op}}_{t,i}(a|\x_t,\r,c)
\end{align}
where the geometric interpolation preserves the non-negativity of the rates by construction. Since the rate decomposes into an operation rate and a token distribution, we apply guidance to each component separately via a \emph{na\"ive rate} CFG as,
\begin{align}
    \hat{\lambda}^{\mathrm{op}}_{t,i}(\x_t,\r,c) &= \lambda^{\mathrm{op}}_{t,i}(\x_t,\r,c)^{1+w} \lambda^{\mathrm{op}}_{t,i}(\x_t,\r)^{-w}  \\
    \tilde{Q}^{\mathrm{op}}_{t,i}(a|\x_t,\r,c) & = Q^{\mathrm{op}}_{t,i}(a|\x_t,\r)^{1-w}Q^{\mathrm{op}}_{t,i}(a|\x_t,\r,c)^w  
\end{align}
where $\hat{\lambda}^{\mathrm{op}}_{t,i}$ scales the operation rate toward the conditioned regime, and $\tilde{Q}^{\mathrm{op}}_{t,i}$ interpolates the token distributions in probability space. However, this guidance operates independently at each token position with no cross-position coherence.
\begin{figure}[!t]
    \centering
    \includegraphics[width=\linewidth]{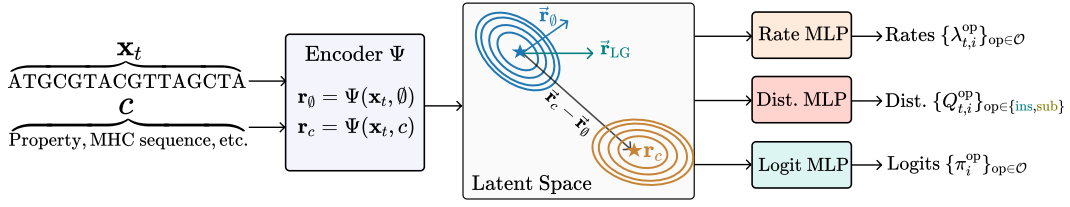}
    \caption{\textbf{Model pipeline with Latent Guidance}. Given a noised sequence $\x_t$ with external conditioning $c$, encoder $\Psi$ produces an unconditional latent $\r_\emptyset$, and a conditional latent $\r_c$, which are interpolated into a guided latent  $\r_{\mathrm{LG}}$ using a guidance scale $w$.This guided latent is then passed into the MLPs to output the edit rates and distributions with auxiliary outputs.}
    \label{fig:latent_guide}
    \vskip -1.0em
\end{figure}
\subsubsection{Latent Guidance} 
A key advantage of our global latent formulation is that it admits a natural, novel form of classifier-free guidance that operates entirely in the continuous latent space, rather than directly on the discrete token rates. Following CFG, during training, we randomly drop the external condition $c$ with probability $p_{\mathrm{drop}}$, training the encoder $\Psi$ jointly to predict both a conditional and unconditional latent
\begin{align}\label{eq:latent}
     \r_{\mathrm{LG}} = \r_\emptyset + w (\r_c - \r_\emptyset), \quad \text{where}~~~\r_c \sim \mathcal{N}(\r; \boldsymbol{\mu}_c, \sigma^2_r \mathrm{I}),~~\text{and}~~\r_\emptyset \sim \mathcal{N}(\r; \boldsymbol{\mu}_\emptyset, \sigma^2_r \mathrm{I}),
\end{align}
where $\boldsymbol{\mu}_\emptyset = \Psi(\mathbf{x}_t, \emptyset)$, $\boldsymbol{\mu}_c = \Psi(\mathbf{x}_t, c)$, $\emptyset$ denotes the null condition and $w \geq 0$ is the guidance scale. At test time, guidance is applied by interpolating between $\r_c$ and $\r_\emptyset$ in the continuous latent space. The guided latent $\r_{\mathrm{LG}}$ then conditions all per-token operations jointly via $p_{\Phi}(\mathrm{op}^i | x^i_t, \r_{\mathrm{LG}})$, propagating the guidance signal globally through the shared $r$, in contrast to rate-space guidance, which steers each position independently. Since it operates in the latent space, non-negativity of the rates is preserved by construction without requiring renormalization. The guided rate is then,
\begin{align*}\label{eq:velocity}
    \tilde{u}^{\theta}_t(\mathbf{x} | \mathbf{x}_t,c) = \mathbb{E}_{\r_{\mathrm{LG}} \sim p_{\Psi}(\r_{\mathrm{LG}} | \mathbf{x}_t,c)}\left[\sum_{i=1}^{N}\sum_{\mathrm{op} \in \mathcal{O}} p_{\Phi}(\mathrm{op}^i = \mathrm{op} | x^i_t, \r_{\mathrm{LG}})\, u^{\theta}_t\left(\mathrm{op}(\mathbf{x}, i) | \mathbf{x}_t, \r_{\mathrm{LG}}\right)\right]
\end{align*}
This latent interpolation admits a principled probabilistic interpretation: the guidance shift corresponds to the score of an implicit latent classifier, as formalized below.
\begin{proposition}[Latent Guidance induces an implicit classifier]\label{prop:latent_guidance}
Let $\r$ be a sufficient statistic for $c$, with unconditional and conditional latents $\r_\emptyset \sim p_\Psi(\r | \mathbf{x}_t) = \mathcal{N}(\r; \boldsymbol{\mu}_\emptyset, \sigma^2_r \mathrm{I})$ and $\r_c \sim p_\Psi(\r | \mathbf{x}_t, c) = \mathcal{N}(\r; \boldsymbol{\mu}_c, \sigma^2_r \mathrm{I})$. Then the guidance direction is proportional to the score of an implicit classifier $p(c | \r)$, and $\r_{\mathrm{LG}}$ amounts to a gradient ascent step on $\log p(c | \r)$ in latent space:
$$
    \r_{\mathrm{LG}} \sim \mathcal{N}(\r_{\mathrm{LG}};\boldsymbol{\mu}_\emptyset + w\sigma^2_r \nabla_\r \log p(c \mid \r)\big|_{\r = \boldsymbol{\mu}_\emptyset}, (1-w)^2\sigma^2_r + w^2\sigma^2_r  \mathrm{I})
$$
\end{proposition}

\definecolor{burntorange}{rgb}{0.8, 0.33, 0.0}

\subsection{Training Objective}
We adopt a training strategy similar to Edit Flows \citep{havasi2025edit}, operating in an augmented space to avoid the intractability of standard cross-entropy or ELBO objectives. Since edit sequences (insertions, deletions, substitutions) induce multiple paths between the same strings, direct transition rates are difficult to compute. Following \citet{havasi2025edit,holderrieth2024generator}, we define the process over an augmented space $\mathcal{X} \times \mathcal{Z}$, enabling a tractable objective via Bregman divergences.

\textbf{Auxiliary alignment process.} Given a pair $(\tilde{\x}_0,\x_1)$ an alignment defines a precise set of edit operations transforming $\tilde{\x}_0$ to $\x_1$ by introducing a blank token $\varepsilon \notin \mathcal{T}$. We define the augmented space $\mathcal{Z} = \mathcal{T} \cup \{ \varepsilon\}$, with $f_{\mathrm{rm-blanks}}:\mathcal{Z} \rightarrow \mathcal{X}$ stripping all $\varepsilon$ tokens. This defines a structured coupling in the augmented space $\pi(\tilde{\z}_0,\z_1,\mathrm{K}_{\gamma})$ satisfying the marginal distributions of the original distributions $p(\tilde{\x}_0|\mathrm{K}_{\gamma})$ and $p_1(\x_1)$. Then, the conditional probability path and the rate is given as,
\begin{align}
    p_t(\x,\z|\tilde{\x}_0,\tilde{\z}_0,\x_1,\z_1) &= p_t(\x,\z|\tilde{\z}_0,\z_1) = p_t(\z|\tilde{\z}_0,\z_1)\delta_{f_\mathrm{rm\text{-}blanks}(\mathbf{z})}(\x) \\
    u_t(\mathbf{x}, \mathbf{z} | \mathbf{x}_t, \mathbf{z}_t, \tilde{\z}_0, \mathbf{z}_1) &= \delta_{f_\mathrm{rm\text{-}blanks}(\mathbf{z})}(\mathbf{x})\sum_{i=1}^{N} \frac{\dot{\kappa}_t}{1-\kappa_t}\left(\delta_{z^i_1}(z^i) - \delta_{z^i_t}(z^i)\right)\delta_{\mathbf{z}_t}(z^{\neg i})
\end{align}
\textbf{Learning Objective.} We learn the parameterized rate $u^{\theta}_t$ by minimizing a Bregman divergence \citep{bregman1967relaxation} between the true and predicted rates over the augmented space, which simplifies to the following objective,
\begin{empheq}[box=\mymath]{equation*}  
\begin{aligned}
&\min_{\theta} \mathcal{L}(\theta) =  
\mathbb{E}_{t,\x_t,\z_t} 
    \left[ \sum_{\x \neq \x_t} \colorA{u^{\theta}_{t}(\x|\x_t)} -
    \sum_{i=1}^N \colorB{\mathbbm{1}_{[z^i_1 \neq z^i_t]} \frac{\dot{\kappa_t}}{1-\kappa_t} \log u^\theta_t(\x(\z_t,i,z_1^i) |\x_t)}
     \right] \\
&~~~~~~~~~~\text{with}~~ 
(\x_t, \z_t) \sim  p_t(\x_t,\z_t|\tilde{\z}_0,\z_1), (\tilde{\z}_0,\z_1) \sim \pi(\tilde{\z}_0,\z_1,\mathrm{K}_{\gamma}),~\text{and}
~~t \sim \mathcal{U}(0,1)
\end{aligned}
\label{eq:lw_new}
\end{empheq}
where $\colorA{u^{\theta}_{t}(\x|\x_t)} = \sum_z \mathbb{E}_{p_t(\tilde{\z}_0,\z_1,\z_t|\x_t)}u_t(\mathbf{x}, \mathbf{z} | \mathbf{x}_t, \mathbf{z}_t, \tilde{\z}_0, \mathbf{z}_1)$ is the marginalized output rate of the model, and $\colorB{\x(\z_t,i,z_1^i)} = f_\mathrm{rm\text{-}blanks}(z^1_t,\ldots,z^{i-1}_t,z^i_t,z^{i+1}_t,\ldots,z^N_t)$  represents a specific edit operation in the sequence space. This loss can be interpreted as minimizing all extraneous output rates of the model while performing the cross-entropy over the edit operations that move $\x_t$ closer to $\x_1$. 
\begin{figure}[!t]
    \includegraphics[width=\linewidth]{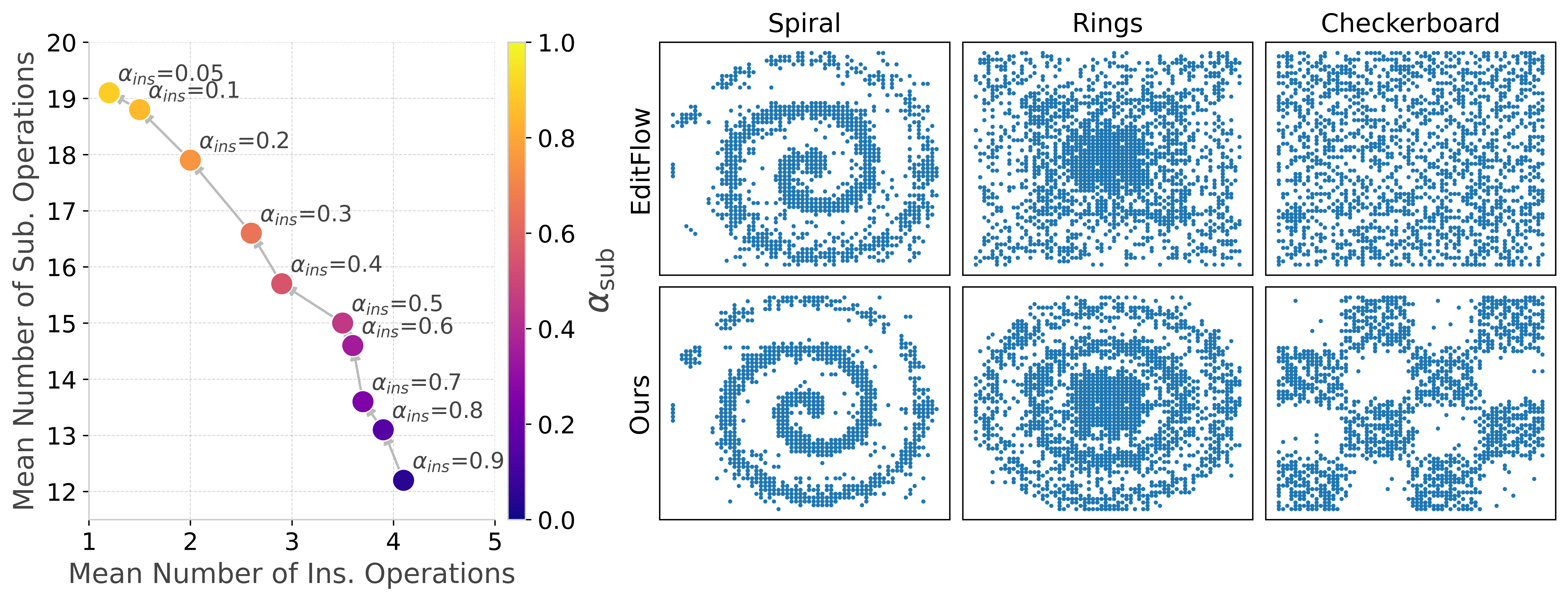}
    \caption{(\emph{\textbf{left}}) Evolution of edit operations during generation as the Dirichlet concentration $\boldsymbol{\alpha}$ is shifted between insertion and substitution vertices (with fixed $\alpha_{\mathrm{del}} = 0.05$), showing model's ability to act as a global budget controller, (\emph{\textbf{right}}) Representative generated distributions on toy datasets.}
    \label{fig:toy_data}
    \vskip -1.5em
\end{figure}
\section{Experiments}

\textbf{Tasks.} We evaluate our method on diverse tasks. In \Cref{sec:toy_data}, we evaluate our method ability to model various densities. In \Cref{sec:enhance} and \Cref{sec:promo}, we assess the methods ability to generate conditional and unconditional DNA sequences, and finally, in \cref{sec:peptide}, we show the performance of our method in generating peptide sequences conditioned on MHC. \looseness=-1

\textbf{Implementation.} Our implementation uses PyTorch, following the architecture and pre-processing steps of the baselines. See App. \ref{sec:implement} for further technical details.  \looseness=-1

\textbf{Baselines.} We compare against a range of diffusion and flow-based methods, including Bit Diffusion, DDSM \citep{albergo2022building}, D3PM \citep{austin2021structured}, Linear FM \citep{lipman2024flow}, Dirichlet FM \citep{stark2024dirichlet}, Dirichlet FDM \citep{huang2026improving}, Gumbel-Softmax FM \citep{tang2025gumbel}, and Fisher FM \citep{davis2024fisher}. We cannot compare to EvoFlows \citep{deutschmann2026evoflows} as there is no code available for it.  \looseness=-1

\subsection{Toy datasets} \label{sec:toy_data}
We first evaluate our method on toy datasets to assess its ability to learn discrete distributional patterns in 1D and 2D, using four canonical patterns: (i) Sine-wave, a 1D sinusoidal pattern testing periodic structure; (ii) Spiral, a 2D curved pattern testing long-range dependencies; (iii) Rings, concentric rings testing multi-modal symmetry; and (iv) Checkerboard, a 2D alternating pattern testing sharp local transitions. We measure distribution quality via the Wasserstein distance between generated and true samples. \Cref{fig:toy_data} and \Cref{tab:pepmhc} describes that our method consistently achieves the lowest $\mathcal{W}_2$ across all patterns and accurately reconstructs each distribution's topological features.  \looseness=-1
\begin{table}[!t]
\caption{(\emph{\textbf{left, top}}) \textbf{Unconditional Enhancer Generation}: Distributional similarity measured via FBD and sampling efficiency (NFE) for $10,000$ generated DNA sequences; $\dagger$ denotes metrics taken from \citet{stark2024dirichlet}. (\emph{\textbf{left, bottom}}) \textbf{Ablation study} on prior choices and structured coupling with respective FBD values over Melanoma and FlyBrain data. (\textit{\textbf{right}}) \textbf{Conditional Promoter Design}: Conditioned on a transcription profile, each method is tasked to generate a DNA sequence with that profile. MSE measures the difference between predicted and ground truth regulatory activities.}
\label{tab:dna}
\begin{minipage}{0.48\textwidth}
\centering
\resizebox{\textwidth}{!}{
\begin{tabular}{l ccc}
    \toprule
    \multirow{2}{*}{Method} & Melanoma & Fly Brain & \multirow{2}{*}{NFE} \\
    \cmidrule{2-3}
    & \multicolumn{2}{c}{Fréchet Bio. Dist. ($\downarrow$)}  & \\
    \midrule
    Random Sequence & $622.8$ & $876$ & $-$ \\
    Language Model$^\dagger$ & $36.0$ & $25.2$ & $500$ \\
    Linear FM$^\dagger$ & $19.6$ & $15.0$ & $100$\\
    Dirichlet FM$^\dagger$ & $5.3$ & $15.2$ & $100$\\
    Fisher FM & $3.8$ & $27.5$ & $100$\\
    EditFlow & $3.7$ & $7.9$ & $100$\\
    \midrule
    \rowcolor{gray!15}
    \textbf{FlexFlow} & $\mathbf{3.4}$ & $\mathbf{4.2}$ & 100 \\
    \bottomrule
\end{tabular}
}
\vskip 0.4em
\resizebox{0.7\textwidth}{!}{
\begin{tabular}{ccccc} 
            \toprule
            $p_{\mathrm{unif}}$ & $p_{\mathrm{freq}}$ & $\mathrm{K}_\gamma$ & MEL & FB \\
            \midrule
            \cmark & & &  3.7 &  5.2 \\
            & \cmark & &  3.5 &  4.7 \\
            & \cmark & \cmark & 3.4 & 4.2  \\
            \bottomrule
        \end{tabular}}
\end{minipage}
\hfill
\begin{minipage}{0.51\textwidth}
\centering
\resizebox{\textwidth}{!}{
\begin{tabular}{l cc}
    \toprule
    Method & MSE  & NFE \\
    \midrule
    Bit Diffusion$^\dagger$ (Bit Encoding) & $0.041$ &  $100$ \\
    Bit Diffusion$^\dagger$(One-hot Encoding) & $0.039$ &  $100$ \\
    D3PM-Uniform$^\dagger$& $0.037$  &  $100$\\
    DDSM$^\dagger$ & $0.033$ &  $100$\\
    Language Model$^\dagger$ & $0.033$ &  $1024$\\
    Linear FM$^\dagger$ & $0.028$ &  $100$\\
    Dirichlet FM$^\dagger$ & $0.026$ & $100$\\
    Dirichlet FDM & $0.026$ &  \textcolor{gray}{(N/A)}\\
    Fisher FM & $0.030$ &  \textcolor{gray}{(N/A)}\\
    Gumbel-Softmax FM & $0.029$ &  \textcolor{gray}{(N/A)}\\
    \midrule
    \rowcolor{gray!15}
    \textbf{FlexFlow} - Rate Guidance & $0.024$ &  $100 $\\
    \rowcolor{gray!15}
    \textbf{FlexFlow} - Latent Guidance w/o $\mathrm{K}_\gamma$ & $0.025$ &  $100 $\\
    \rowcolor{gray!15}
    \textbf{FlexFlow} - Latent Guidance & $\textbf{0.022}$ &  $100 $\\
    \bottomrule
\end{tabular}}

\end{minipage}
\end{table}
\subsection{Enhancer DNA Sequence Design}\label{sec:enhance}
\textbf{Data.} We evaluate on two enhancer sequence datasets derived from fly brain cells \citep{janssens2022decoding} and human melanoma cells \citep{atak2021interpretation}. These consist of 104k and 89k sequences of length 500 with ATAC-seq measurements \citep{buenrostro2013transposition} across 81 and 47 classes.

\textbf{Metrics and Superior performance in unconditional design.} To assess the similarity between the data and model distributions, we use the Fréchet Biological Distance (FBD) introduced by \citet{stark2024dirichlet}. This involves training a classifier to predict cell types and using its hidden representations as embeddings to compute Wasserstein distance for both generated and real samples. \Cref{tab:dna} shows the results on both enhancer datasets. Notably, our method is able to achieve the lowest FBD as compared to the baselines, demonstrating the enhanced fidelity afforded by our method. 

\subsection{Promoter DNA Sequence Design}\label{sec:promo}
\textbf{Data.} We next evaluate the conditional design of DNA promoter sequences given a desired promoter profile, following the setup of Dirichlet FM \citep{stark2024dirichlet} and DDSM \citep{avdeyev2023dirichlet}. We use 100,000 human promoter sequences of 1024 base pairs from \citet{hon2017atlas}, each annotated with a FANTOM5 CAGE signal \citep{shiraki2003cap,fantom2014promoter} representing per-position transcription initiation probability ($\mathbf{r} \in \mathbb{R}^{1024}$). Chromosomes 8 and 9 are held out for testing; the rest is used for training.

\textbf{Metrics and Superior performance in conditional design.} Following \citet{stark2024dirichlet}, we evaluate generated sequences using the MSE between their predicted regulatory activity and that of the corresponding original sequence. Regulatory activity is estimated using the promoter-specific outputs of SEI model \citep{chen2022sequence}. We report the regulatory activity prediction results in \Cref{tab:dna}. Our method significantly outperforms both Dirichlet FM $(0.026)$ and DDSM $(0.033)$, achieving a lower mean squared error (MSE) across the test set. While our model using Rate Guidance already surpasses current flows ($0.024$), Latent Guidance provides the best performance. 

\subsection{MHC Peptide Sequence Design}\label{sec:peptide}
We consider a conditional generation task where, given an MHC allele $m$, the goal is to generate a short peptide $\mathbf{x}$ (typically of length 13--18 amino acids) that is naturally processed and presented by the given MHC molecule. The dataset thus consists of pairs $\{(m_i, \mathbf{x}_i)\}_{i=1}^{N}$ (see Appendix~\ref{app:dataset}).

\textbf{Data.} Using Eluted Ligand (EL) data from \cite{nilssonAccuratePredictionHLA2023}, we adapted a peptide--MHC~II benchmark \citep{korpela2026pepgen}, using peptides physically presented on MHC molecules, collected via mass spectrometry. The data is partitioned into $\mathcal{D}_{\mathrm{Train}}$ and $\mathcal{D}_{\mathrm{Test}}$ using a strict sequence-based split ensuring no 9-mer subsequence is shared. For more details see App.~\ref{app:dataset} 

\textbf{Metrics.} We evaluate peptide generation with two complementary metrics. (i) \textbf{Discriminator score} ($\in [0,1]$): a held-out DeepMHCII model \citep{youDeepMHCIINovelBinding2022} retrained on EL data (Appendix~\ref{app:discriminator}) to separate real ligands from synthetic negatives, with higher scores indicating closer resemblance to true ligands. (ii) \textbf{Fréchet Protein Distance} (FPD): the Wasserstein distance between ESM2 \citep{lin2023evolutionary} embeddings of generated and test peptides, where lower is better. We treat the discriminator score as our primary metric, since it directly rewards presentation plausibility in a setting where only positive ligands are observed; FPD serves as a complementary check on distributional fidelity.
\begin{table}[!t]
\caption{(\textit{\textbf{left}}) \textbf{Evaluation on test dataset for MHC Peptide Sequence Design}: FPD, DeepMHCII classifier score values, and NFE on $10,000$ generated peptide sequence for held out test set $\mathcal{D}_{\mathrm{Test}}$ of MHC sequences. (\textit{\textbf{right}}) \textbf{Wasserstein distance $\mathcal{W}_2$ between true and generated distribution.} }
\label{tab:pepmhc}
\begin{minipage}{0.57\textwidth}
\centering
\resizebox{\textwidth}{!}{
\begin{tabular}{l ccc}
    \toprule
    Method & FPD $\downarrow$  & DeepMHCII $\uparrow$ & Uniq. \\
    \midrule
    Dirichlet FM & $0.79$ & $0.21$ & $100$\\
    Dirichlet FDM & $0.72$ & $0.31$ & $100$\\
    Fisher FM & $0.75$ & $0.28$& $100$ \\
    \midrule
    \rowcolor{gray!15}
    \textbf{FlexFlow} - Rate Guidance & \textbf{0.58} & 0.58 &  $100 $\\
    \rowcolor{gray!15}
    \textbf{FlexFlow} - Latent Guidance & 0.84 & \textbf{0.66} & $100 $\\
    \bottomrule
\end{tabular}}
\end{minipage}
\hfill
\begin{minipage}{0.425\textwidth}
\centering
\resizebox{\textwidth}{!}{
\begin{tabular}{l cc>{\columncolor{gray!15}}c}
    \toprule
    Data & D3PM & EditFlow & \textbf{FlexFlow} \\
    \midrule
    Sine wave & $0.033$ & $0.031$ & $\mathbf{0.024}$ \\
    Spiral & $0.056$ & $0.045$ & $\mathbf{0.042}$ \\
    Rings & $0.085$ & $0.080$ & $\mathbf{0.040}$ \\
    Checkerboard & $0.067$ & $0.070$ & $\mathbf{0.041}$ \\
    \bottomrule
\end{tabular}}
\end{minipage}
\vskip -1.0em
\end{table}

\textbf{Superior results in peptide sequence generation.} Our model substantially 
outperforms all baselines on DeepMHCII score --- Dirichlet FM (0.21), Dirichlet 
FDM (0.31), and Fisher FM (0.28) --- reaching 0.58 with rate-space guidance and 
0.66 with latent-space guidance (Table~\ref{tab:pepmhc}). Rate-space guidance also 
improves FPD over all baselines, while latent-space guidance prioritizes binding 
plausibility at the cost of distributional coverage --- a tradeoff we ablate in 
the next section.

\section{Ablation Studies}
\textbf{Scalability of the Global Latent Representation.} We conducted an ablation study to assess how the latent encoder scales with context length by measuring its peak GPU memory and forward-pass latency. \Cref{tab:abl} (left) shows that both remain nearly flat up to 1k residues and grow only modestly up to 10k. This follows from the dilated CNN encoder, which scales linearly in sequence length and captures long-range context without the quadratic cost of self-attention, making it well suited to longer genomic contexts.

\textbf{Biological Viability of Forced Edit Budgets.} We conducted an ablation study to evaluate the biological plausibility of sequences generated under skewed edit budgets, varying the Dirichlet concentration parameters $\boldsymbol{\alpha}$ on the peptide sequence design dataset. \Cref{tab:abl} (right) shows that substitution-biased budgets, which more closely reflect the typical substitution-heavy setting, yield the best FPD and DeepMHCII scores, while insertion or deletion biases lead to lower performance.

\textbf{Effect of $\sigma_r$.} We conducted an ablation study varying $\sigma_r$ and evaluating downstream biological plausibility via the DeepMHCII score. \Cref{tab:abl} (middle) shows that performance is fairly stable across values, with a dip at $\sigma_r = 0$, indicating that a non-zero $\sigma_r$ improves downstream performance.

\begin{table}[!t]
    \centering
    \caption{\textbf{Ablation studies.} (\textbf{\emph{left}}) Encoder scalability with context length, (\textbf{\emph{center}}) effect of $\sigma_r$ on downstream plausibility, and (\textbf{\emph{right}}) biological viability under forced edit budgets.}
    \label{tab:abl}
    \resizebox{0.38\linewidth}{!}{\begin{tabular}{lcc}
        \toprule
        Length &  Latency (ms) & Peak Memory (GB) \\
        \midrule
        100	& 3.14&	0.093 \\
        500	&3.46&	0.095 \\
        1024	&3.61&	0.098 \\
        2000	&6.21	&0.108 \\
        10000	&24.31&	0.614 \\
        \bottomrule
    \end{tabular}} 
    \hfill
    \resizebox{0.2\linewidth}{!}{\begin{tabular}{lc}
        \toprule
        $\sigma_r$ &  DeepMHCII ($\uparrow$) \\
        \midrule
        0.0 &0.62 \\
        0.1 &0.66 \\
        0.5 &0.64 \\
        \bottomrule
    \end{tabular}} 
    \hfill
    \resizebox{0.38\linewidth}{!}{\begin{tabular}{lcccc}
        \toprule
        $\alpha_{\mathrm{ins}}$&$\alpha_{\mathrm{sub}}$ & $\alpha_{\mathrm{del}}$ & FPD &DeepMHCII  \\
        \midrule
        0.9 & 0.05 & 0.05 & 1.05 & 0.22\\
        0.05 & 0.9 & 0.05 & 0.58 & 0.58\\
        0.05 & 0.05 & 0.9 & 0.98 & 0.22\\
        \bottomrule
    \end{tabular}} 
    \vskip -1.5em
\end{table}

\textbf{Post-hoc Operation Control.} We conducted an ablation study to show the post-hoc operation control over the edit operations. \Cref{fig:toy_data} illustrates that shifting concentration parameters $\boldsymbol{\alpha}$ toward specific simplex vertices prioritizes corresponding edit operations. This allows $\boldsymbol{\alpha}$ to act as a budget controller, tuning operation frequencies during inference to meet specific requirements without retraining.

\textbf{Time complexity.} We compare inference speed against prior methods by measuring the time to generate a 500-length DNA sequence over 100 reverse timesteps. \Cref{tab:abl_time} in the appendix shows that our method achieves a comparable inference speed compared to the baselines.

\textbf{Choice of prior and the effect of structured coupling.} We conduct an ablation study to evaluate the impact of the prior and the structured coupling on the fidelity of generated DNA sequences. \Cref{tab:dna} shows that the frequency-informed prior $p_{\mathrm{freq}}$ with structured coupling $\mathrm{K}_\gamma$ gives peak performance by aligning the generative path with domain-specific biases. 

\begin{figure}[!hbt]
    \centering
    \includegraphics[width=\linewidth]{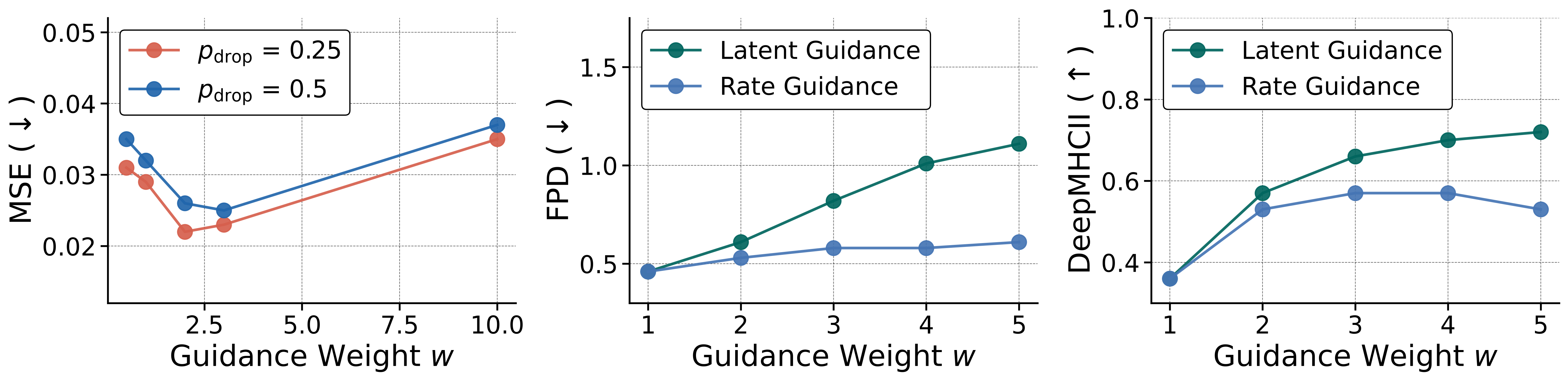}
    \caption{\textbf{Diversity vs Quality tradeoff and effect of $p_{\mathrm{drop}}$}. (\emph{left}) Impact of $p_{\mathrm{drop}}$ on promoter DNA sequence design. (\emph{center, right}) At high guidance scales, LG produces higher-quality peptide ligands than RG at the cost of increased FPD, illustrating a diversity-quality tradeoff.}
    \label{fig:scale}
    \vskip -1em
\end{figure}

\textbf{Guidance Choice: Effects of Scale and $p_{\mathrm{drop}}$ on Diversity -- Quality trade-off.} We conducted an ablation study to characterize the impact of different guidance strategies and the interplay between the guidance scale $w$ and $p_{\mathrm{drop}}$. \Cref{tab:dna,tab:pepmhc} compare various formulations, while \Cref{fig:scale} shows that higher $p_{\mathrm{drop}}$ needs a larger $w$ to maintain fidelity, shifting the optimal performance point accordingly.

\section{Conclusion}

We introduce \textbf{FlexFlow}, a Discrete Flow Matching framework for biological sequence design that addresses key limitations of existing discrete generative methods through a biologically informed structured coupling, an edit-based reverse flow parameterization with Dirichlet-controlled operation mixing for flexible-length generation, and a latent guidance scheme that preserves global structural consistency. Across unconditional and conditional DNA generation, as well as peptide design on our newly introduced peptide-MHC II benchmark, FlexFlow consistently outperforms strong diffusion and flow-matching baselines. Future work includes adapting distillation-based training objectives to FlexFlow,  enabling high-quality sequence generation in just a handful of sampling steps.

\section*{Acknowledgements}
YV and VG acknowledge support from the Research Council of Finland for the “Human-steered next-generation machine learning for reviving drug design” project (grant decision 342077). YV thanks Suomen Tekniikan edistämissäätiö (grant number 10477) for their support. VG also acknowledges support from the Jane and Aatos Erkko Foundation (grant 7001703) for “Biodesign: Use of artificial intelligence in enzyme design for synthetic biology”. We acknowledge generous computational support from the Aalto-IT Science project and ELLIS Institute Finland. YV thanks Priscilla Ong for fruitful discussions during the initial ideation phase and the CNY retreat in Singapore.
\bibliography{ref}

@article{lipman2024flow,
  title={Flow matching guide and code},
  author={Lipman, Yaron and Havasi, Marton and Holderrieth, Peter and Shaul, Neta and Le, Matt and Karrer, Brian and Chen, Ricky TQ and Lopez-Paz, David and Ben-Hamu, Heli and Gat, Itai},
  journal={arXiv preprint arXiv:2412.06264},
  year={2024}
}

@article{shi2024simplified,
  title={Simplified and generalized masked diffusion for discrete data},
  author={Shi, Jiaxin and Han, Kehang and Wang, Zhe and Doucet, Arnaud and Titsias, Michalis},
  journal={Advances in neural information processing systems},
  volume={37},
  pages={103131--103167},
  year={2024}
}

@article{korpela2026pepgen,
  title={PepGen: conditional generation of peptides for MHC binding},
  author={Korpela, Dani and Dumitrescu, Alexandru and Strazar, Martin and Li, Rui and Xavier, Ramnik J and Graham, Daniel B and L{\"a}hdesm{\"a}ki, Harri},
  journal={Bioinformatics},
  volume={42},
  number={Supplement\_2},
  pages={btag464},
  year={2026},
  publisher={Oxford University Press}
}

@article{kong2025protflow,
  title={Protflow: fast protein sequence design via flow matching on compressed protein language model embeddings},
  author={Kong, Zitai and Zhu, Yiheng and Xu, Yinlong and Zhou, Hanjing and Yin, Mingzhe and Wu, Jialu and Xu, Hongxia and Hsieh, Chang-Yu and Hou, Tingjun and Wu, Jian},
  journal={arXiv preprint arXiv:2504.10983},
  year={2025}
}

@inproceedings{
tang2025peptune,
title={PepTune: De Novo Generation of Therapeutic Peptides with Multi-Objective-Guided Discrete Diffusion},
author={Sophia Tang and Yinuo Zhang and Pranam Chatterjee},
booktitle={Frontiers in Probabilistic Inference: Learning meets Sampling},
year={2025},
url={https://openreview.net/forum?id=eBoJ9YRx0w}
}

@article{sahoo2024simple,
  title={Simple and effective masked diffusion language models},
  author={Sahoo, Subham S and Arriola, Marianne and Schiff, Yair and Gokaslan, Aaron and Marroquin, Edgar and Chiu, Justin T and Rush, Alexander and Kuleshov, Volodymyr},
  journal={Advances in Neural Information Processing Systems},
  volume={37},
  pages={130136--130184},
  year={2024}
}

@article{zhou2023dnabert,
  title={Dnabert-2: Efficient foundation model and benchmark for multi-species genome},
  author={Zhou, Zhihan and Ji, Yanrong and Li, Weijian and Dutta, Pratik and Davuluri, Ramana and Liu, Han},
  journal={arXiv preprint arXiv:2306.15006},
  year={2023}
}

@article{truong2025understanding,
  title={Understanding protein function with a multimodal retrieval-augmented foundation model},
  author={Truong Jr, Timothy Fei and Bepler, Tristan},
  journal={arXiv preprint arXiv:2508.04724},
  year={2025}
}

@article{truong2023poet,
  title={Poet: A generative model of protein families as sequences-of-sequences},
  author={Truong Jr, Timothy and Bepler, Tristan},
  journal={Advances in Neural Information Processing Systems},
  volume={36},
  pages={77379--77415},
  year={2023}
}

@article{deutschmann2026evoflows,
  title={EvoFlows: Evolutionary Edit-Based Flow-Matching for Protein Engineering},
  author={Deutschmann, Nicolas and Ferragu, Constance and Ziegler, Jonathan D and Aziznejad, Shayan and Bixby, Eli},
  journal={arXiv preprint arXiv:2603.11703},
  year={2026}
}

@inproceedings{verma2023abode,
  title={Abode: Ab initio antibody design using conjoined odes},
  author={Verma, Yogesh and Heinonen, Markus and Garg, Vikas},
  booktitle={International Conference on Machine Learning},
  pages={35037--35050},
  year={2023},
  organization={PMLR}
}

@article{henikoff1992amino,
  title={Amino acid substitution matrices from protein blocks.},
  author={Henikoff, Steven and Henikoff, Jorja G},
  journal={Proceedings of the national academy of sciences},
  volume={89},
  number={22},
  pages={10915--10919},
  year={1992}
}

@article{lin2023evolutionary,
  title={Evolutionary-scale prediction of atomic-level protein structure with a language model},
  author={Lin, Zeming and Akin, Halil and Rao, Roshan and Hie, Brian and Zhu, Zhongkai and Lu, Wenting and Smetanin, Nikita and Verkuil, Robert and Kabeli, Ori and Shmueli, Yaniv and others},
  journal={Science},
  volume={379},
  number={6637},
  pages={1123--1130},
  year={2023},
  publisher={American Association for the Advancement of Science}
}

@article{hayes2025simulating,
  title={Simulating 500 million years of evolution with a language model},
  author={Hayes, Thomas and Rao, Roshan and Akin, Halil and Sofroniew, Nicholas J and Oktay, Deniz and Lin, Zeming and Verkuil, Robert and Tran, Vincent Q and Deaton, Jonathan and Wiggert, Marius and others},
  journal={Science},
  volume={387},
  number={6736},
  pages={850--858},
  year={2025},
  publisher={American Association for the Advancement of Science}
}

@article{alamdari2023protein,
  title={Protein generation with evolutionary diffusion: sequence is all you need},
  author={Alamdari, Sarah and Thakkar, Nitya and Van Den Berg, Rianne and Tenenholtz, Neil and Strome, Robert and Moses, Alan M and Lu, Alex X and Fusi, Nicol{\`o} and Amini, Ava P and Yang, Kevin K},
  journal={BioRxiv},
  pages={2023--09},
  year={2023},
  publisher={Cold Spring Harbor Laboratory}
}

@inproceedings{avdeyev2023dirichlet,
  title={Dirichlet diffusion score model for biological sequence generation},
  author={Avdeyev, Pavel and Shi, Chenlai and Tan, Yuhao and Dudnyk, Kseniia and Zhou, Jian},
  booktitle={International Conference on Machine Learning},
  pages={1276--1301},
  year={2023},
  organization={PMLR}
}

@article{nijkamp2023progen2,
  title={Progen2: exploring the boundaries of protein language models},
  author={Nijkamp, Erik and Ruffolo, Jeffrey A and Weinstein, Eli N and Naik, Nikhil and Madani, Ali},
  journal={Cell systems},
  volume={14},
  number={11},
  pages={968--978},
  year={2023},
  publisher={Elsevier}
}

@article{albergo2022building,
  title={Building normalizing flows with stochastic interpolants},
  author={Albergo, Michael S and Vanden-Eijnden, Eric},
  journal={arXiv preprint arXiv:2209.15571},
  year={2022}
}

@article{ferruz2022protgpt2,
  title={ProtGPT2 is a deep unsupervised language model for protein design},
  author={Ferruz, Noelia and Schmidt, Steffen and H{\"o}cker, Birte},
  journal={Nature communications},
  volume={13},
  number={1},
  pages={4348},
  year={2022},
  publisher={Nature Publishing Group UK London}
}

@article{austin2021structured,
  title={Structured denoising diffusion models in discrete state-spaces},
  author={Austin, Jacob and Johnson, Daniel D and Ho, Jonathan and Tarlow, Daniel and Van Den Berg, Rianne},
  journal={Advances in neural information processing systems},
  volume={34},
  pages={17981--17993},
  year={2021}
}

@article{tang2025gumbel,
  title={Gumbel-softmax flow matching with straight-through guidance for controllable biological sequence generation},
  author={Tang, Sophia and Zhang, Yinuo and Tong, Alexander and Chatterjee, Pranam},
  journal={ArXiv},
  pages={arXiv--2503},
  year={2025}
}

@article{davis2024fisher,
  title={Fisher flow matching for generative modeling over discrete data},
  author={Davis, Oscar and Kessler, Samuel and Petrache, Mircea and Ceylan, {\.I}smail {\.I} and Bronstein, Michael and Bose, Avishek J},
  journal={Advances in Neural Information Processing Systems},
  volume={37},
  pages={139054--139084},
  year={2024}
}

@article{chen2022sequence,
  title={A sequence-based global map of regulatory activity for deciphering human genetics},
  author={Chen, Kathleen M and Wong, Aaron K and Troyanskaya, Olga G and Zhou, Jian},
  journal={Nature genetics},
  volume={54},
  number={7},
  pages={940--949},
  year={2022},
  publisher={Nature Publishing Group US New York}
}

@article{huang2026improving,
  title={Improving flow matching by aligning flow divergence},
  author={Huang, Yuhao and Transue, Taos and Wang, Shih-Hsin and Feldman, William and Zhang, Hong and Wang, Bao},
  journal={arXiv preprint arXiv:2602.00869},
  year={2026}
}

@article{fantom2014promoter,
  title={A promoter-level mammalian expression atlas},
  author={The FANTOM Consortium and the RIKEN PMI and CLST (DGT)},
  journal={Nature},
  volume={507},
  number={7493},
  pages={462--470},
  year={2014},
  publisher={Nature Publishing Group UK London}
}

@article{shiraki2003cap,
  title={Cap analysis gene expression for high-throughput analysis of transcriptional starting point and identification of promoter usage},
  author={Shiraki, Toshiyuki and Kondo, Shinji and Katayama, Shintaro and Waki, Kazunori and Kasukawa, Takeya and Kawaji, Hideya and Kodzius, Rimantas and Watahiki, Akira and Nakamura, Mari and Arakawa, Takahiro and others},
  journal={Proceedings of the National Academy of Sciences},
  volume={100},
  number={26},
  pages={15776--15781},
  year={2003},
  publisher={National Academy of Sciences}
}

@article{hon2017atlas,
  title={An atlas of human long non-coding RNAs with accurate 5' ends},
  author={Hon, Chung-Chau and Ramilowski, Jordan A and Harshbarger, Jayson and Bertin, Nicolas and Rackham, Owen JL and Gough, Julian and Denisenko, Elena and Schmeier, Sebastian and Poulsen, Thomas M and Severin, Jessica and others},
  journal={Nature},
  volume={543},
  number={7644},
  pages={199--204},
  year={2017},
  publisher={Nature Publishing Group UK London}
}

@article{buenrostro2013transposition,
  title={Transposition of native chromatin for fast and sensitive epigenomic profiling of open chromatin, DNA-binding proteins and nucleosome position},
  author={Buenrostro, Jason D and Giresi, Paul G and Zaba, Lisa C and Chang, Howard Y and Greenleaf, William J},
  journal={Nature methods},
  volume={10},
  number={12},
  pages={1213--1218},
  year={2013},
  publisher={Nature Publishing Group US New York}
}

@article{atak2021interpretation,
  title={Interpretation of allele-specific chromatin accessibility using cell state--aware deep learning},
  author={Atak, Zeynep Kalender and Taskiran, Ibrahim Ihsan and Demeulemeester, Jonas and Flerin, Christopher and Mauduit, David and Minnoye, Liesbeth and Hulselmans, Gert and Christiaens, Valerie and Ghanem, Ghanem-Elias and Wouters, Jasper and others},
  journal={Genome research},
  volume={31},
  number={6},
  pages={1082--1096},
  year={2021},
  publisher={Cold Spring Harbor Lab}
}

@article{janssens2022decoding,
  title={Decoding gene regulation in the fly brain},
  author={Janssens, Jasper and Aibar, Sara and Taskiran, Ibrahim Ihsan and Ismail, Joy N and Gomez, Alicia Estacio and Aughey, Gabriel and Spanier, Katina I and De Rop, Florian V and Gonzalez-Blas, Carmen Bravo and Dionne, Marc and others},
  journal={Nature},
  volume={601},
  number={7894},
  pages={630--636},
  year={2022},
  publisher={Nature Publishing Group UK London}
}

@inproceedings{weilbach2023graphically,
  title={Graphically structured diffusion models},
  author={Weilbach, Christian Dietrich and Harvey, William and Wood, Frank},
  booktitle={International Conference on Machine Learning},
  pages={36887--36909},
  year={2023},
  organization={PMLR}
}

@article{lou2023discrete,
  title={Discrete diffusion modeling by estimating the ratios of the data distribution},
  author={Lou, Aaron and Meng, Chenlin and Ermon, Stefano},
  journal={arXiv preprint arXiv:2310.16834},
  year={2023}
}

@article{nisonoff2024unlocking,
  title={Unlocking guidance for discrete state-space diffusion and flow models},
  author={Nisonoff, Hunter and Xiong, Junhao and Allenspach, Stephan and Listgarten, Jennifer},
  journal={arXiv preprint arXiv:2406.01572},
  year={2024}
}

@article{ho2022classifier,
  title={Classifier-free diffusion guidance},
  author={Ho, Jonathan and Salimans, Tim},
  journal={arXiv preprint arXiv:2207.12598},
  year={2022}
}

@article{alido2025whitened,
  title={Whitened Score Diffusion: A Structured Prior for Imaging Inverse Problems},
  author={Alido, Jeffrey and Li, Tongyu and Sun, Yu and Tian, Lei},
  journal={arXiv preprint arXiv:2505.10311},
  year={2025}
}

@article{stark2024dirichlet,
  title={Dirichlet flow matching with applications to dna sequence design},
  author={Stark, Hannes and Jing, Bowen and Wang, Chenyu and Corso, Gabriele and Berger, Bonnie and Barzilay, Regina and Jaakkola, Tommi},
  journal={arXiv preprint arXiv:2402.05841},
  year={2024}
}

@article{shaul2024flow,
  title={Flow matching with general discrete paths: A kinetic-optimal perspective},
  author={Shaul, Neta and Gat, Itai and Havasi, Marton and Severo, Daniel and Sriram, Anuroop and Holderrieth, Peter and Karrer, Brian and Lipman, Yaron and Chen, Ricky TQ},
  journal={arXiv preprint arXiv:2412.03487},
  year={2024}
}

@article{holderrieth2024generator,
  title={Generator matching: Generative modeling with arbitrary markov processes},
  author={Holderrieth, Peter and Havasi, Marton and Yim, Jason and Shaul, Neta and Gat, Itai and Jaakkola, Tommi and Karrer, Brian and Chen, Ricky TQ and Lipman, Yaron},
  journal={arXiv preprint arXiv:2410.20587},
  year={2024}
}

@article{bregman1967relaxation,
  title={The relaxation method of finding the common point of convex sets and its application to the solution of problems in convex programming},
  author={Bregman, Lev M},
  journal={USSR computational mathematics and mathematical physics},
  volume={7},
  number={3},
  pages={200--217},
  year={1967},
  publisher={Elsevier}
}

@article{campbell2024generative,
  title={Generative flows on discrete state-spaces: Enabling multimodal flows with applications to protein co-design},
  author={Campbell, Andrew and Yim, Jason and Barzilay, Regina and Rainforth, Tom and Jaakkola, Tommi},
  journal={arXiv preprint arXiv:2402.04997},
  year={2024}
}

@article{jukes1969evolution,
  title={Evolution of protein molecules},
  author={Jukes, Thomas H and Cantor, Charles R and others},
  journal={Mammalian protein metabolism},
  volume={3},
  number={21},
  pages={132},
  year={1969},
  publisher={New York}
}

@article{hasegawa1985dating,
  title={Dating of the human-ape splitting by a molecular clock of mitochondrial DNA},
  author={Hasegawa, Masami and Kishino, Hirohisa and Yano, Taka-aki},
  journal={Journal of molecular evolution},
  volume={22},
  number={2},
  pages={160--174},
  year={1985},
  publisher={Springer}
}

@article{gat2024discrete,
  title={Discrete flow matching},
  author={Gat, Itai and Remez, Tal and Shaul, Neta and Kreuk, Felix and Chen, Ricky TQ and Synnaeve, Gabriel and Adi, Yossi and Lipman, Yaron},
  journal={Advances in Neural Information Processing Systems},
  volume={37},
  pages={133345--133385},
  year={2024}
}

@article{havasi2025edit,
  title={Edit Flows: Flow Matching with Edit Operations},
  author={Havasi, Marton and Karrer, Brian and Gat, Itai and Chen, Ricky TQ},
  journal={arXiv preprint arXiv:2506.09018},
  year={2025}
}

@article{youDeepMHCIINovelBinding2022,
  title = {{{DeepMHCII}}: A Novel Binding Core-Aware Deep Interaction Model for Accurate {{MHC-II}} Peptide Binding Affinity Prediction},
  shorttitle = {{{DeepMHCII}}},
  author = {You, Ronghui and Qu, Wei and Mamitsuka, Hiroshi and Zhu, Shanfeng},
  year = 2022,
  month = jun,
  journal = {Bioinformatics},
  volume = {38},
  number = {Supplement\_1},
  pages = {i220-i228},
  publisher = {Oxford Academic},
  issn = {1367-4803},
  doi = {},
  urldate = {2026-01-27},
  langid = {english},
}

@article{nilssonAccuratePredictionHLA2023,
  title = {Accurate Prediction of {{HLA}} Class {{II}} Antigen Presentation across All Loci Using Tailored Data Acquisition and Refined Machine Learning},
  author = {Nilsson, Jonas B. and Kaabinejadian, Saghar and Yari, Hooman and Kester, Michel G.~D. and {van Balen}, Peter and Hildebrand, William H. and Nielsen, Morten},
  year = 2023,
  month = nov,
  journal = {Science Advances},
  volume = {9},
  number = {47},
  pages = {},
  publisher = {American Association for the Advancement of Science},
}

\newpage
\appendix

\section{Proofs}\label{sec:proof}

\subsection{Proof of \cref{prop:coupling}}
For any $\x^i_0 \in \x_0 \in \mathcal{T}$, 
\begin{align}
    \mathrm{K}_{\gamma}(\tilde{\x}^i_0|\x^i_0)=\mathrm{K}_{\gamma}\e^{i}_{\x_0} = \frac{1}{|\mathcal{T}|}\mathbf{1}\mathbf{1}^\top \e^{i}_{\x_0} = \frac{1}{|\mathcal{T}|}\mathbf{1}
\end{align}
providing,
\begin{align}
    \mathrm{Cat}(\tilde{x}^{i}_{0}; \mathrm{K}_{\gamma} \e^{i}_{\x_0}) = \mathrm{Uniform}(\mathcal{T})
\end{align}
regardless of $\x^i_0$. The marginalization over $\x_0$ trivially factors out, giving 
\begin{align}
    q(\tilde{\x}_0 | \mathrm{K}_{\gamma}) = \mathrm{Uniform}(\mathcal{T}^N)
\end{align}
providing the coupling
\begin{align}
        \pi(\tilde{\x}_0, \mathbf{x}_1) = \mathrm{Uniform}(\mathcal{T}^N)\,p_{\mathrm{data}}(\mathbf{x}_1)
\end{align}
\subsection{Proof of \cref{prop:latent_guidance}}

Let $p(\r|\x_t) = \mathcal{N}(\r;\boldsymbol{\mu}_\emptyset,\sigma^2_r \mathrm{I})$, and $p(\r|\x_t,c) = \mathcal{N}(\r;\boldsymbol{\mu}_c,\sigma^2_r \mathrm{I})$. Then,
\begin{align}
    \r_{\mathrm{LG}} = \r_\emptyset + w(\r_c - \r_\emptyset) \sim \mathcal{N}(\boldsymbol{\mu}_\emptyset + w(\boldsymbol{\mu}_c - \boldsymbol{\mu}_\emptyset), (1-w)^2\sigma^2_r + w^2\sigma^2_r  \mathrm{I})
\end{align}
By utilizing Bayes' theorem and assuming $\r$ is a sufficient statistic, we get,
\begin{align}
    p(\r \mid \x_t,c) &\propto p(c\mid \r) p(\r \mid \x_t) \\
    \nabla_\r \log p(\r \mid \x_t,c) &= \nabla_\r \log p(\r \mid \x_t) + \nabla_\r \log p(c \mid \r) \\
    \frac{\boldsymbol{\mu}_c - \r}{\sigma^2_r} &= \frac{\boldsymbol{\mu}_\emptyset - \r}{\sigma^2_r} + \nabla_\r \log p(c \mid \r) \\
    \boldsymbol{\mu}_c - \boldsymbol{\mu}_\emptyset &= \sigma^2_r \nabla_\r \log p(c \mid \r)
\end{align}
This provides the latent guidance as,
\begin{align}
    \r_{\mathrm{LG}} \sim \mathcal{N}(\boldsymbol{\mu}_\emptyset + w\sigma^2_r \nabla_\r \log p(c \mid \r)\big|_{\r = \boldsymbol{\mu}_\emptyset}, (1-w)^2\sigma^2_r + w^2\sigma^2_r  \mathrm{I})
\end{align}

\section{Dataset Construction}
\label{app:dataset}

MHC-II molecules are heterodimeric proteins composed of an $\alpha$- and a 
$\beta$-chain, with binding specificity jointly determined by both chains. 
Rather than encoding full protein sequences, ML pipelines typically represent 
each MHC-II molecule via a \textit{pseudosequence} --- a 34-residue subset of 
amino acid positions most critical for peptide binding interactions \cite{nilssonAccuratePredictionHLA2023}.
The MHC-II pseudosequences (conditionings) are used as input features 
 for generation of their corresponding peptide sequences (targets) throughout our pipeline.

\begin{figure}
    \centering
    \includegraphics[width=\linewidth]{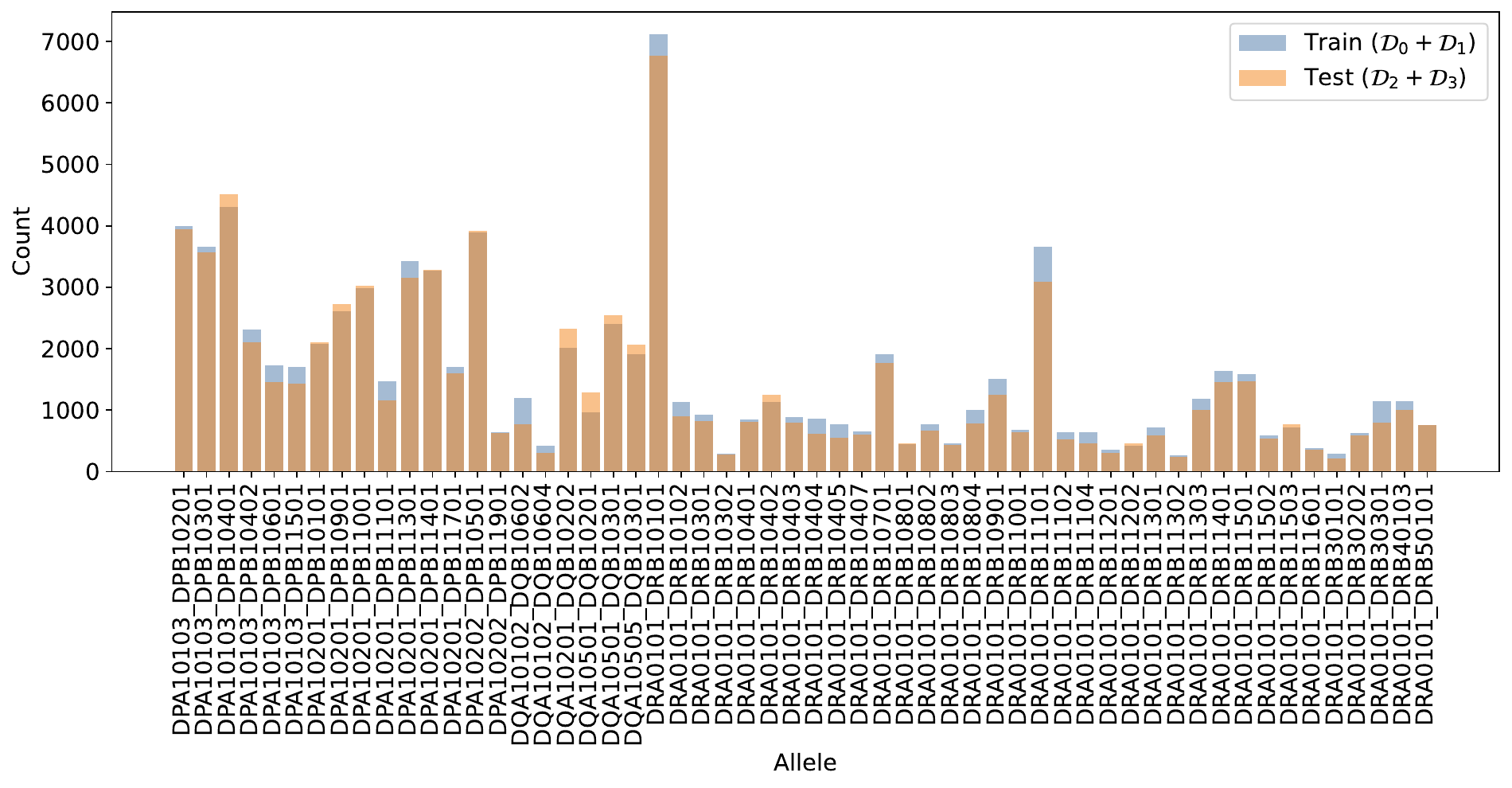}
    \caption{Allele distribution of train and test sets for the peptide MHC binding task}
    \label{fig:allele}
\end{figure}

The dataset comprises peptides distributed across 56 MHC-II allele combinations 
($\alpha$/$\beta$ chain pairs) from the three canonical allele groups of 
HLA class II: DR, DP, and DQ. The dataset exhibits significant class imbalance, largely reflecting differences 
in experimental data availability across alleles, with per-allele combination 
counts ranging from 1{,}518 (\texttt{DRA*01:01\_DRB5*01:01}) to 13{,}892 
(\texttt{DRA*01:01\_DRB1*01:01}) peptides across the combined train and test sets. The per-allele distribution across splits is shown \cref{fig:allele}.

We used the eluted ligand (EL) subset from \cite{nilssonAccuratePredictionHLA2023} and removed all non-standard amino acids (e.g.\ selenocysteine). The original work splits data per allele, ensuring no shared 9-mer among peptides of the same allele within a fold. We applied a stricter criterion: all peptide sequences were clustered jointly across alleles such that no 9-mer is shared between any two clusters. Clusters were then joined to form two balanced partitions, $\mathcal{D}_{\mathrm{Train}}$ and $\mathcal{D}_{\mathrm{Test}}$, each further divided into training and validation subsets as in \cite{korpela2026pepgen}:
\[
    \mathcal{D}_{\mathrm{Train}} = \mathcal{D}_{0} \cup \mathcal{D}_{1}, \qquad
    \mathcal{D}_{\mathrm{Test}}  = \mathcal{D}_{2} \cup \mathcal{D}_{3}.
\]
The generative model was trained on $\mathcal{D}_{0}$ and validated on $\mathcal{D}_{1}$. The test discriminator was trained on $\mathcal{D}_{2}$ and validated on $\mathcal{D}_{3}$ (see Appendix~\ref{app:discriminator}). The synthetic negatives from \cite{nilssonAccuratePredictionHLA2023} were used only for discriminator training and not for generative model training. We evaluated the FPD between $\mathcal{D}_1$ and $\mathcal{D}_3$ and between $\mathcal{D}_3$ and corresponding artificial negatives for reference yielding 0.30 and 1.98, respectively.

\section{Implementation Details}\label{sec:implement}

\begin{table}[!h]
    \centering
    \caption{Inference speed and Trainable Parameters for DNA sequence design.}
    \label{tab:abl_time}
    \resizebox{0.7\linewidth}{!}{\begin{tabular}{llcc}
        \toprule
        Method &  Model size & Trainable params  & Inference time (sec.) \\
        \midrule
        Dirichlet FM &$\sim$ 5M & $\sim$ 5M  & $\sim$ 3.8 \\
        Fisher FM &$\sim$ 5M & $\sim$ 5M  & $\sim$ 4.5 \\
        Dirichlet FDM &$\sim$ 5M & $\sim$ 5M  & $\sim$ 4.3 \\
        FlexFlow & $\sim$ 5M & $\sim$ 5M & $\sim$ 4.0 \\
        \bottomrule
    \end{tabular}} 
\end{table}

\subsection{DNA Sequence Design}
The model was trained on a single NVIDIA V100 GPU.
\begin{table}[!hbt]
\caption{Default hyperparameters for DNA sequence design experiment.}
\label{tab:hyper_model}
\begin{center}
\resizebox{0.95\textwidth}{!}{
\begin{tabular}{lrlc}
\toprule
Experiment & Hyperparameter & Meaning & Value \\
\midrule
\multirow{6}{*}{Enhancer Design} &Hidden dimension & Hidden dimension & $256$ \\
&CNN stacks & Number of CNN blocks& $4$ \\
&CNN layers & Number of CNN layers in one stack& $5$ \\
&Kernel size & Size of filters in a stack & $[9,9,9,9,9] $ \\
&Padding & Padding in a stack & $[4,4,16,64,256] $   \\
&Dilation & Dilation in a stack & $[0,0,4,16,64] $   \\
\midrule
\multirow{8}{*}{Promoter Design} &Hidden dimension & Hidden dimension & $256$ \\
&CNN stacks & Number of CNN blocks& $4$ \\
&CNN layers & Number of CNN layers in one stack& $5$ \\
&Kernel size & Size of filters in a stack & $[9,9,9,9,9] $ \\
&Padding & Padding in a stack & $[4,4,16,64,256] $   \\
&Dilation & Dilation in a stack & $[0,0,4,16,64] $   \\
&$p_{\mathrm{drop}}$ & Unconditional drop rate & $0.5 $   \\
&$w$ & Guidance scale for CFG/LG & $3 $   \\
\midrule
\multirow{6}{*}{Training} &Max epochs & Total number of Epochs & $1000$ \\
&lr & Learning Rate& $5e-4$ \\
&Optimizer & Optimizer & Adam \\
&Batch Size & Samples in a batch & 256  \\
&Reverse time steps & Integration time steps & 100 \\
&Prior & Prior distribution& $p_{\mathrm{freq}}$ \\
\bottomrule
\end{tabular}}
\end{center}
\end{table}

\subsection{MHC Peptide Sequence Design}
\label{app:mhcpep}

The model was trained with a ddp strategy on four NVIDIA V100 GPUs.

\begin{table}[!hbt]
\caption{Default hyperparameters for Peptide sequence design experiment.}
\label{tab:hyper_model}
\begin{center}
\resizebox{0.95\textwidth}{!}{
\begin{tabular}{lrlc}
\toprule
Experiment & Hyperparameter & Meaning & Value \\
\midrule
\multirow{6}{*}{Peptide Design} &Hidden dimension & Hidden dimension & $128$ \\
&CNN stacks & Number of CNN blocks& $2$ \\
&CNN layers & Number of CNN layers in one stack& $4$ \\
&Kernel size & Size of filters in a stack & $[9,9,9,9,9] $ \\
&Padding & Padding in a stack & $[4,4,16,64,256] $   \\
&Dilation & Dilation in a stack & $[0,0,4,16,64] $   \\
&$p_{\mathrm{drop}}$ & Unconditional drop rate & $0.2 $   \\
&$w$ & Guidance scale for CFG/LG & $3 $   \\
&Dropout &Dropout after each conv layer& 0.3\\
& Gaussian std& Latent std $\sigma_r$& 0.1\\
\midrule
\multirow{6}{*}{Training} &Max epochs & Limit of number of Epochs & $3000$ \\
&Early stopping&Early stopping patience& 1000\\
&lr & Learning Rate& $5e-4$ \\
&Optimizer & Optimizer & Adam \\
&Weight decay& Weight Decay Regularization&$1e-4$\\
&lr scheduler& Scheduler for Learning Rate& cosine\\
&Batch Size & Samples in a batch & 256  \\
&Reverse time steps & Integration time steps & 100 \\
&Prior & Prior distribution& $p_{\mathrm{freq}}$ \\
\bottomrule
\end{tabular}}
\end{center}
\end{table}
\subsubsection{Test Discriminator Training}
\label{app:discriminator}

We adapted DeepMHCII \citep{youDeepMHCIINovelBinding2022} to serve as a held-out discriminator for evaluating generated peptides. The original DeepMHCII model was developed for binding affinity prediction (regression); we repurposed it for binary eluted ligand classification by replacing the mean squared error loss with binary cross-entropy. No architectural changes were required, as DeepMHCII's convolutional kernels already capture core flanking regions---amino acids outside the MHC binding groove---which are particularly informative for EL data compared to affinity data.

The discriminator was trained on $\mathcal{D}_{2}$ and validated on $\mathcal{D}_{3}$, using the positive EL pairs from \cite{nilssonAccuratePredictionHLA2023} together with synthetic negatives subsampled to a 5:1 negative-to-positive ratio. To verify discriminator quality, we evaluated it on $\mathcal{D}_{1}$ (unseen during discriminator training) paired with negatives from \cite{nilssonAccuratePredictionHLA2023}, achieving an AUROC of 0.971, indicating strong discriminative performance.

To support validation of the generative model without data leakage, we additionally trained a second discriminator on $\mathcal{D}_{\mathrm{Train}}$ following the same procedure.

\section{Constructing $\mathrm{K}_{\gamma}$ matrices} \label{sec:k_gamma}
We provide a reference implementation for constructing $\mathrm{K}_\gamma$ for proteins from a BLOSUM substitution matrix. The procedure converts BLOSUM log-odds scores into a row-stochastic transition kernel. 

\begin{lstlisting}[language=Python]
import torch
from sklearn.preprocessing import normalize
import itertools
from collections import Counter, OrderedDict
import csv
import pandas as pd
import subprocess
import os
import urllib
import numpy as np

STOP = '*'
GAP = '-'
MSA_PAD = '!'
MASK = '#' 
START = '@'
OTHER_AAS = 'JOU'
AAINDEX_ALPHABET = 'ARNDCQEGHILKMFPSTWYV'
AMB_AAS = 'BZX'
MSA_AAS = ALL_AAS + GAP
MSA_ALPHABET = ALL_AAS + GAP + STOP + MASK + START + MSA_PAD
BLOSUM_AAS = AAINDEX_ALPHABET + AMB_AAS
BLOSUM_ALPHABET = BLOSUM_AAS + OTHER_AAS + GAP + MSA_PAD + STOP + MASK + START
BLOSUM_EXTRAS = 'JOU-'


def loadMatrix(path):
    """
    Taken from https://pypi.org/project/blosum/
    Edited slightly from original implementation

    Reads a Blosum matrix from file. Changed slightly to read in larger blosum matrix
    File in a format like:
        https://www.ncbi.nlm.nih.gov/IEB/ToolBox/C_DOC/lxr/source/data/BLOSUM62
    Input:
        path: str, path to a file.
    Returns:
        blosumDict: Dictionary, The blosum dict
    """

    with open(path, "r") as f:
        content = f.readlines()

    blosumDict = {}

    header = True
    for line in content:
        line = line.strip()

        # Skip comments starting with #
        if line.startswith(";"):
            continue

        linelist = line.split()

        # Extract labels only once
        if header:
            labelslist = linelist
            header = False
            
            continue

        if not len(linelist) == len(labelslist) + 1:
            print(len(linelist), len(labelslist))
            # Check if line has as may entries as labels
            raise EOFError("Blosum file is missing values.")

        for index, lab in enumerate(labelslist, start=1):
            blosumDict[f"{linelist[0]}{lab}"] = float(linelist[index])

    if not len(blosumDict) == len(labelslist) ** 2:
        print(len(blosumDict), len(labelslist))
        raise EOFError("Blosum file is not quadratic.", len(blosumDict), len(labelslist)**2)
    return blosumDict

def softmax(x):
    """
    Compute softmax over x
    """
    return np.exp(x)/np.sum(np.exp(x),axis=0)

def double_stochastic(q):
    q_norm = normalize(q, axis=1, norm='l1')
    while not np.isclose(np.min(np.sum(q_norm, axis=0)), 1): # only checking that one value converges to 1 (prob best to do all 4 min/max)
        q_norm = normalize(q_norm, axis=0, norm='l1')
        q_norm = normalize(q_norm, axis=1, norm='l1')
    return q_norm


def K_blosum(matrix):
        alphabet = list("".join(MSA_ALPHABET))
        all_aas = list("".join(MSA_AAS))
        a_to_i = {u: i for i, u in enumerate(alphabet)}

        q = np.array([i for i in matrix.values()])
        q = q.reshape((len(all_aas),len(all_aas)))
        q = softmax(q)
        q = double_stochastic(q)
        q = torch.tensor(q)
        # REORDER BLOSUM MATRIX BASED ON MSA_ALPHABET (self.alphabet, self.a_to_i)
        new_q = q.clone()
        i2_to_a = np.array(list(BLOSUM_ALPHABET))
        for i, row in enumerate(new_q):
            for j, value in enumerate(row):
                ind1, ind2 = [i, j]
                key = i2_to_a[ind1], i2_to_a[ind2]
                new1, new2 = [a_to_i[k] for k in key]
                new_q[new1, new2] = q[ind1, ind2]

        return new_q
\end{lstlisting}


\end{document}